\documentclass[10pt,journal,compsoc,twoside]{IEEEtran}
\usepackage{amsfonts,bm,booktabs,graphicx,hyperref,multirow,textcomp,url}
\usepackage{algorithm,algpseudocode}
\usepackage{setspace}
\usepackage{subfigure}
\usepackage{algpseudocode}
\usepackage{amsmath}
\usepackage{mathrsfs}
\usepackage{float}
\usepackage{color}
\usepackage{tikz}
\usepackage{comment}
\usepackage{marvosym}
\usepackage{caption}
\usepackage{paralist}
\usepackage{enumitem}
\usepackage{colortbl}

\makeatletter 
\renewcommand{\@thesubfigure}{\hskip\subfiglabelskip}
\newcommand{\etal}{{\emph{et al.~}}\@ }

\ifCLASSOPTIONcompsoc
  \usepackage[nocompress]{cite}
\else
  \usepackage{cite}
\fi
\ifCLASSINFOpdf
\else
\fi

\begin{document}
\title{Exploiting Optical Flow Guidance for Transformer-Based Video Inpainting}

\author{Kaidong Zhang,
        Jialun Peng,
        Jingjing Fu,~\IEEEmembership{Member,~IEEE,}
        and Dong Liu,~\IEEEmembership{Senior~Member,~IEEE}% <-this % stops a space
\IEEEcompsocitemizethanks{\IEEEcompsocthanksitem Date of current version \today. This work was supported by the Natural Science Foundation of China under Grant 62036005, and by the Fundamental Research Funds for the Central Universities under Grant WK3490000006. \emph{(Corresponding author: Dong Liu.)}
\IEEEcompsocthanksitem K. Zhang, J. Peng, and D. Liu are with the MOE Key Laboratory of Brain-Inspired Intelligent Perception and Cognition, University of Science and Technology of China, Hefei 230027, China (e-mail: richu@mail.ustc.edu.cn; pjl@mail.ustc.edu.cn; dongeliu@ustc.edu.cn).
\IEEEcompsocthanksitem J. Fu is with Microsoft Research Asia, Beijing 100190, China (e-mail: jifu@microsoft.com).
}}% <-this % stops a space
% \thanks{Manuscript received April 19, 2005; revised August 26, 2015.}}


\markboth{IEEE Transactions on Pattern Analysis and Machine Intelligence}%
{Zhang \MakeLowercase{\textit{et al.}}: Exploiting Optical Flow Guidance for Transformer-Based Video Inpainting}



\IEEEtitleabstractindextext{%
\begin{abstract}
Transformers have been widely used for video processing owing to the multi-head self attention (MHSA) mechanism. However, the MHSA mechanism encounters an intrinsic difficulty for video inpainting, since the features associated with the corrupted regions are degraded and incur inaccurate self attention. This problem, termed query degradation, may be mitigated by first completing optical flows and then using the flows to guide the self attention, which was verified in our previous work -- flow-guided transformer (FGT). We further exploit the flow guidance and propose FGT++ to pursue more effective and efficient video inpainting. First, we design a lightweight flow completion network by using local aggregation and edge loss. Second, to address the query degradation, we propose a flow guidance feature integration module, which uses the motion discrepancy to enhance the features, together with a flow-guided feature propagation module that warps the features according to the flows. Third, we decouple the transformer along the temporal and spatial dimensions, where flows are used to select the tokens through a temporally deformable MHSA mechanism, and global tokens are combined with the inner-window local tokens through a dual-perspective MHSA mechanism. FGT++ is experimentally evaluated to be outperforming the existing video inpainting networks qualitatively and quantitatively.
\end{abstract}

}

\maketitle

% is not selected <OR> if conference mode is selected - because compsoc
% conference papers position the abstract like regular (non-compsoc)
\IEEEdisplaynontitleabstractindextext

% \fi
\IEEEpeerreviewmaketitle

\ifCLASSOPTIONcompsoc
\IEEEraisesectionheading{\section{Introduction}\label{sec:introduction}}
\else
\section{Introduction}
\label{sec:introduction}
\fi

\IEEEPARstart{V}{ideo} inpainting, aiming at filling in the corrupted regions in videos with plausible content \cite{bertalmio2001navier}, has been widely applied in object removal \cite{10.1111/j.1467-8659.2012.03000.x}, video retargeting \cite{kim2019deep}, video stabilization \cite{1634345}, and so on. Ideal video inpainting should maintain the spatiotemporal coherence in the completed videos, so that the inpainted regions are as imperceptible as possible. Such a goal is challenging since it requires accurate modeling along both spatial and temporal dimensions. Compared with image inpainting~\cite{pathakCVPR16context,IizukaSIGGRAPH2017,liu2018partialinpainting,yu2018generative,yu2018free,Nazeri_2019_ICCV,9113276,peng2021generating,Liao_2021_CVPR}, video inpainting concerns more about the temporal dimension. If using image inpainting to video frames individually, the results are seldom satisfactory since they lack the temporal consistency perceived in natural videos.

\begin{figure}[t]
\captionsetup{font={scriptsize}}
    \centering
    \subfigure[(a) Input]{\begin{minipage}[t]{0.24\linewidth}
        \centering\includegraphics[width=1\linewidth]{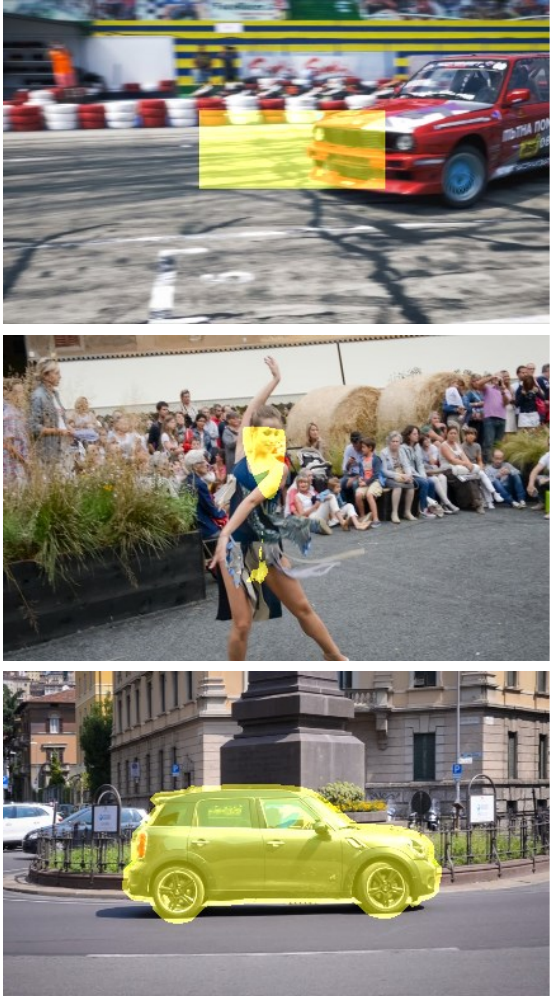}\end{minipage}}
    \subfigure[(b) FGT~\cite{zhang2022flow}]{\begin{minipage}[t]{0.24\linewidth}
        \centering\includegraphics[width=1\linewidth]{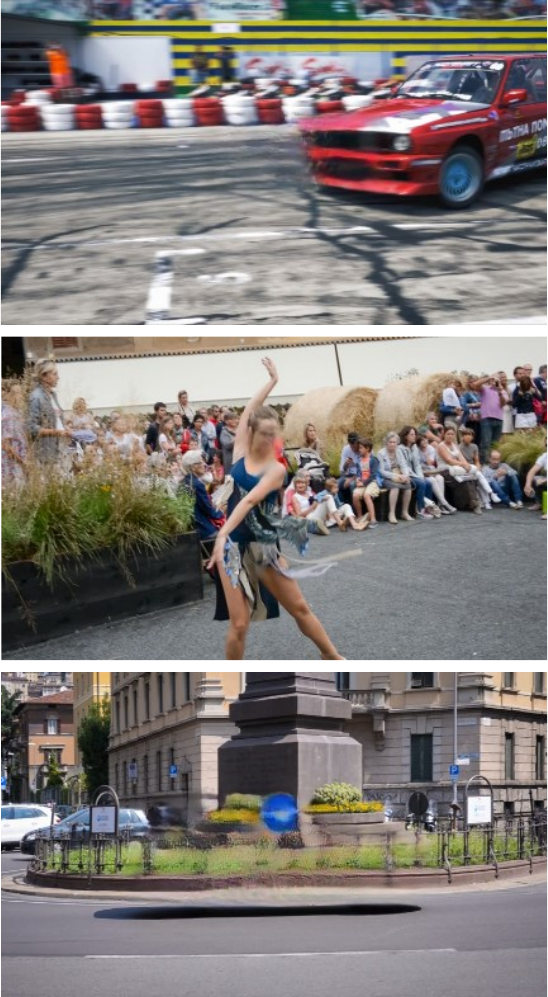}\end{minipage}}
    \subfigure[(c) E2FGVI~\cite{liCvpr22vInpainting}]{\begin{minipage}[t]{0.24\linewidth}
        \centering\includegraphics[width=1\linewidth]{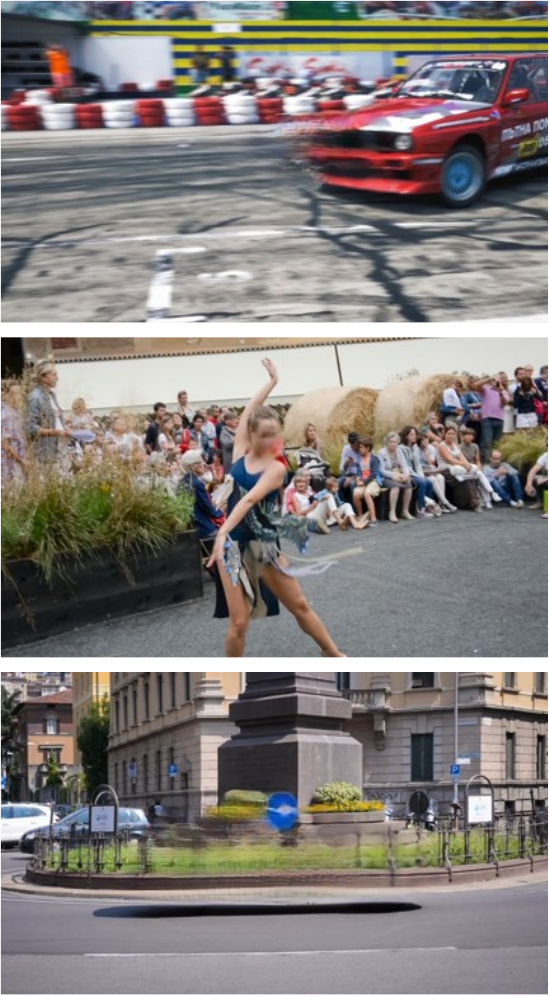}\end{minipage}}
    \subfigure[(d) FGT++]{\begin{minipage}[t]{0.24\linewidth}
        \centering\includegraphics[width=1\linewidth]{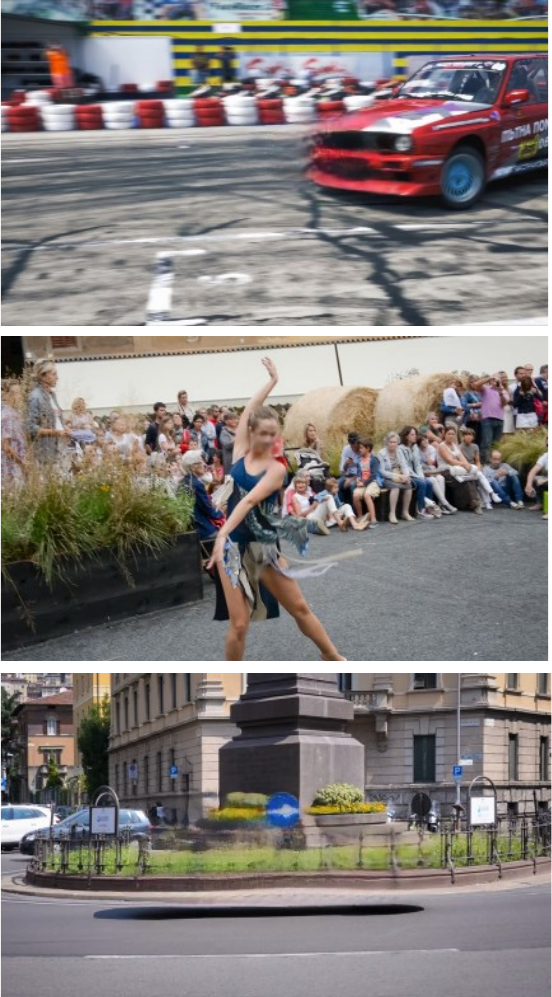}\end{minipage}}
    \captionsetup{font={normalsize}}
    \caption{Qualitative comparison between FGT~\cite{zhang2022flow}, E2FGVI~\cite{liCvpr22vInpainting}, and FGT++. FGT++ is capable of synthesizing more complete scene structure and finer details, which leads to better spatiotemporal coherence in video inpainting.}
    \label{fig:teaser}
\end{figure}

Recently, transformer~\cite{vaswani2017attention} has been used for video inpainting~\cite{yan2020sttn,Liu_2021_FuseFormer,liu2021decoupled,liCvpr22vInpainting} due to its remarkable long-term spatiotemporal modeling ability. In most of the existing studies of transformer-based video inpainting, the multiple video frames are individually encoded into tokens. The feature relevance between the tokens associated with the corrupted regions and the valid regions is estimated by various transformer blocks. Next, the relevance is used to aggregate the features to obtain the features of the corrupted regions, and the obtained features are used to synthesize the inpainted frames. Transformer-based methods have achieved great success and outperformed their previous rivals. However, there is an intrinsic difficulty in adopting transformers for video inpainting. The transformers are built upon the multi-head self attention (MHSA) mechanism, which is the key technology for estimating feature relevance and restoring the features of the corrupted regions. Since the features associated with the corrupted regions are inaccurate (they cannot be accurate because of the frame-wise token generation). These inaccurate features incur error-prone estimation of feature relevance, which further leads to unsatisfactory restored features. This problem is named query degradation in this paper. Indeed, similar problems exist in other transformer-based image/video processing tasks, such as denoising~\cite{10.1007/978-3-031-19800-7_28,liang2022vrt} and super-resolution~\cite{liang2021swinir,liu2022learning}, but for inpainting, the problem is the most typical as the features are non-uniformly degraded: the features associated with the corrupted regions are degraded the most, while the features from the valid regions are not degraded at all.

One solution to the query degradation problem is to first complete the optical flows of the corrupted video and then use the flows to guide the self attention in the transformers. This solution is reasonable because: First, it is easier to complete the flows than to complete the frames since the former are more regular (e.g. well approximated by piecewise smooth signal)~\cite{Xu_2019_CVPR}; Second, the completed flows serve as a strong indicator for spatiotemporal coherence, so we can utilize the flow guidance to retrieve correct tokens with a degraded query in transformers. Indeed, this solution had been experimentally verified in our previous work, named flow-guided transformer (FGT) \cite{zhang2022flow}. FGT had demonstrated its potential but still had several limitations. First, FGT is not so much a solution to the query degradation problem as a workaround for the problem. Second, the flow-guided transformer architecture was not fully investigated in FGT. Therefore, how to exploit the completed optical flows in transformer-based video inpainting is worthy of in-depth investigation.

In this paper, we extend our previous work FGT \cite{zhang2022flow} and conduct a comprehensive study to exploit the flow guidance for transformer-based video inpainting. We propose FGT++ as a more effective video inpainting method that maintains computational efficiency as possible. Following FGT, FGT++ is composed of two networks, a flow completion network and a flow-guided transformer.

For the flow completion network, we notice the fact that the motion fields are likely to be correlated in a temporally local window due to inertia, and we propose to aggregate the features of local flows to use their complementary nature, which greatly improves the flow completion accuracy over the previous studies \cite{Xu_2019_CVPR,Gao-ECCV-FGVC}. For a decent performance-complexity tradeoff, we use the spatial-temporal-decoupled pseudo 3D (P3D) blocks \cite{qiu2017learning} to build a U-Net-like \cite{ronneberger2015u} encoder. We also introduce a new edge loss when training the flow completion network, which helps reconstruct sharper edges in the completed flows, i.e., sharper motion boundaries, without any additional inference cost.

We explicitly address the query degradation problem by proposing two orthogonal modules.
First, we propose a \textbf{F}low \textbf{G}uidance \textbf{F}eature \textbf{I}ntegration (FGFI) module, which utilizes the completed optical flows to supply motion discrepancy to the encoded features. Second, we propose a \textbf{F}low-\textbf{G}uided \textbf{F}eature \textbf{P}ropagation (FGFP) module, which propagates the features along the temporal dimension based on the trajectories suggested by the completed flows, before performing MHSA in the transformer units. To deal with possible errors in the completed flows, we further use deformable convolution \cite{zhu2019deformable} to predict offsets to refine the trajectories. In this manner, we ameliorate the feature quality in the corrupted regions based on the features exposed in nearby frames.

We redesign the flow-guided transformer in FGT to incorporate the proposed FGFI and FGFP modules. Specifically, we decouple the transformer along the temporal and spatial dimensions, i.e., the transformer consists of temporal transformer units and spatial transformer units. In both temporal and spatial units, we use the window partition strategy \cite{Liu_2021_ICCV,yang2021focal,chu2021Twins} to improve efficiency without sacrificing effectiveness. In the temporal transformers, in addition to the large-window attention to ensure enough receptive field, we propose a temporally deformable MHSA (TD-MHSA) mechanism, which uses the completed flows to select the tokens in a much smaller window. In the spatial transformers, we restrict the attention within a small window to pursue less computational cost, which nonetheless limits the receptive field; so we further condense the tokens from the entire token map and integrate such global tokens with the inner-window local tokens through a dual perspective MHSA (DP-MHSA) mechanism.

In the training of the flow-guided transformer, we introduce an amplitude loss, i.e., the Fourier spectrum amplitude differences between ground-truth and inpainted video frames. We demonstrate that the amplitude loss is effective for refining the low-frequency content in the inpainted videos. To our knowledge, such Fourier spectrum losses were not studied for inpainting in the literature.

In summary, the contributions we have made in this paper include:
\begin{itemize}
    \item We analyze the query degradation problem in transformer-based video inpainting, and propose FGFI and FGFP modules to mitigate the problem.
    \item We propose a flow-guided transformer architecture, including TD-MHSA and DP-MHSA mechanisms for temporal and spatial transformer units, respectively.
    \item We design a flow completion network with local flow feature aggregation, which outperforms previous flow completion networks significantly.
    \item We conduct extensive experiments to demonstrate the effectiveness and efficiency of the proposed methods. Our FGT++ is superior to previous video inpainting networks qualitatively and quantitatively, as shown in Fig.~\ref{fig:teaser}.
\end{itemize}

\begin{figure*}[t]
\begin{center}
\includegraphics[width=\linewidth]{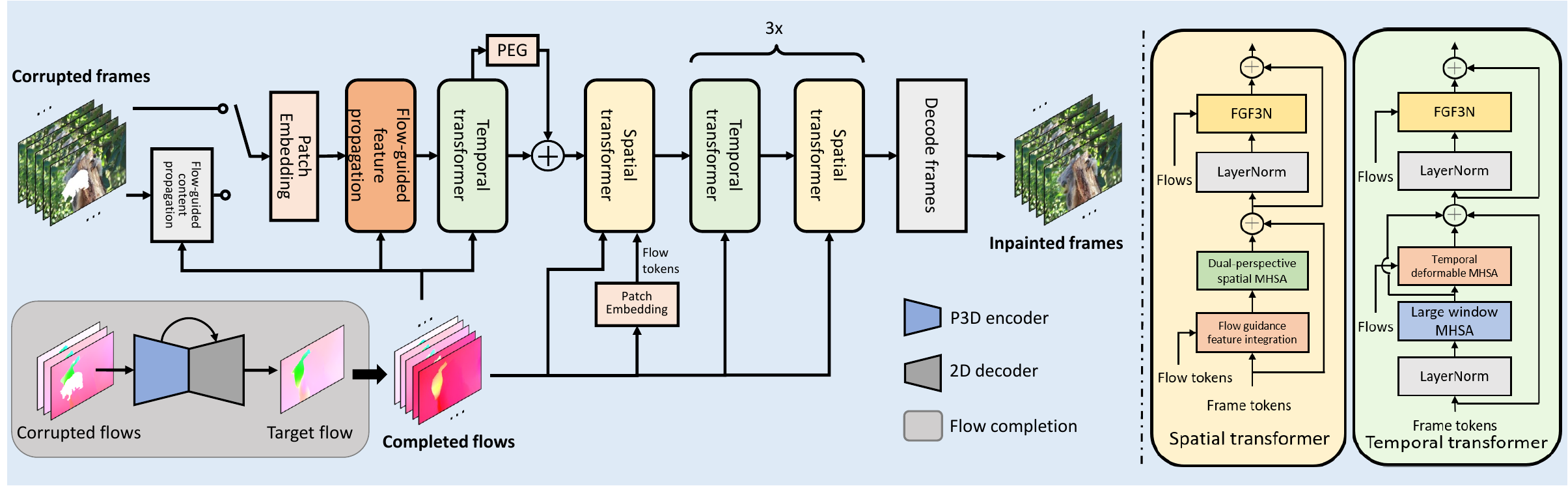}
\end{center}
   \caption{Our method consists of two steps. Firstly, we adopt the \textbf{L}ocal \textbf{A}ggregation \textbf{F}low \textbf{C}ompletion (LAFC) network to complete the corrupted flows. Secondly, we synthesize the corrupted regions with the improved flow-guided transformer under the guidance of the completed optical flows. The ``Flow-guided content propagation" module is optional. PEG: Position embedding generator.}
\label{fig:pipeline}
\end{figure*}

\section{Related Work}\label{Relatedwork}
\subsection{Video Inpainting}
Traditional video inpainting methods \cite{bertalmio2001navier,10.1111/j.1467-8659.2012.03000.x,1634345,7112116,granados2012b,newson2014video} adopt homography or optical flows to explore the geometry relationship through the whole video to propagate the content from valid regions to invalid regions spatiotemporally. Specifically, Huang \etal \cite{Huang-SigAsia-2016} exploit the intrinsic property of natural videos and design a set of energy equations for joint optimizing the motion field and frame quality iteratively, which achieves coherent video inpainting quality with unprecedented fidelity.

Recently, researchers put more effort into deep learning-based video inpainting, which can be divided into flow-based methods and pixel-oriented methods. Flow-based methods \cite{Xu_2019_CVPR,Gao-ECCV-FGVC,Zhang_2022_CVPR} firstly complete optical flows and then utilize the completed flows to capture the correspondence between the valid regions and the corrupted regions in a chain-like manner through all video frames. Our method also includes the flow completion component. Besides the frame content propagation, we explore the usage of completed optical flows in feature propagation, feature enhancement, and temporal attention guidance for better video completion quality.

The second category targets on directly synthesizing the corrupted regions in video frames with the help of spatiotemporal context. Some works adopt 3D CNN \cite{wang2019video,chang2019free}  or channel shift \cite{chang2019learnable,zou2020progressive,ke2021voin} to facilitate the interaction between complementary features in a local temporal window. Several methods integrate recurrent \cite{kim2019deep,Li_2020} or attention \cite{lee2019cpnet,Oh_2019_ICCV} mechanism to CNN-based networks, which is beneficial for extending the limited receptive field of traditional convolution. Inspired by the spatiotemporal redundancy in videos, Zhang \etal \cite{zhang2019internal} and Ouyang \etal \cite{ouyang2021video} adopt internal learning to perform long-range propagation for video inpainting. Currently, Zeng \etal \cite{yan2020sttn}, Liu \etal \cite{Liu_2021_FuseFormer,liu2021decoupled} and Li \etal \cite{liCvpr22vInpainting} adapt transformer \cite{vaswani2017attention} to retrieve similar features in a considerable temporal receptive field for high-quality video inpainting. Our method is also built on the transformer, but differently, we explicitly exploit the motion correspondence across frames under the guidance of completed optical flows to ease the query degradation problem and provide temporal relevance prior to temporal transformer blocks.

\subsection{Image Inpainting}
Image inpainting aims at filling the corrupted regions in images with plausible content to maintain the spatial coherence of the completed images. Traditional Image inpainting methods can be divided into two categories, \textit{i.e.} diffusion-based methods and patch-based methods. The seminal work of Bertalmio \etal \cite{10.1145/344779.344972} proposes an image inpainting problem for the first time, which uses PDE to progressively propagate content from the boundary to the corrupted regions. As a representative work in patch-based image inpainting, PatchMatch \cite{Barnes:2009:PAR} adopts a fast nearest-neighbor field algorithm to generate high-quality inpainting results and reduces computational cost simultaneously. Recently, deep learning-based image inpainting methods utilize the powerful semantic analysis ability of CNN and GAN \cite{goodfellow2020generative} to synthesize new content that may not exist in the corrupted images \cite{pathakCVPR16context, IizukaSIGGRAPH2017, yu2018generative}. Partial convolution \cite{liu2018partialinpainting} and gated convolution \cite{yu2018free} are proposed to fill the free-form holes to complement the limitations of the traditional CNN. Recently, researchers introduced the structure guidance \cite{Nazeri_2019_ICCV, 9113276} and the semantic guidance \cite{Liao_2021_CVPR} to further improve the performance of image inpainting. Different from image inpainting, video inpainting requires coherence not only in one frame, but across all the frames through a video, which is more challenging.

\subsection{Transformers in Computer Vision} 
Recently, transformer \cite{vaswani2017attention} sparked the computer vision community due to its outstanding long-range feature capture ability. Transformer has been integrated to numerous fields and achieved promising performance, such as basic architecture design \cite{Liu_2021_ICCV,yang2021focal,chu2021Twins}, image classification \cite{dosovitskiy2021an,Bhojanapalli_2021_ICCV,Fan_2021_ICCV,Wu_2021_ICCV}, object detection \cite{carion2020end,Misra_2021_ICCV}, action detection \cite{Wang_2021_ICCV}, segmentation \cite{wang2021max}, etc. Our method is an advanced adaptation of the transformer in video inpainting from the perspective of motion exploitation. Besides, we also propose several meticulously designed strategies to improve efficiency while maintaining competitive performance, including spatial-temporal decomposition, temporal deformable MHSA, and the combination of local and global tokes in spatial MHSA.

\section{Method}\label{method}
% \subsection{Problem Formulation}\label{formulation}
Assume the video length is $T$, the input of video inpainting is a corrupted video sequence $X:=$\{$X_1, ..., X_T$\} and its corresponding mask sequence $M:=$\{$M_1,...,M_T$\}. In each mask $M_t$,  ``1" indicates the missing regions, and ``0" represents the valid regions. Our goal is to synthesize the missing regions in the video sequence and maintain the spatiotemporal coherence between our result $\hat{Y}:=$\{$\hat{Y}_1, ..., \hat{Y}_T$\} and the ground truth video sequence $Y:=$\{$Y_1,..., Y_T$\}. 

% \subsection{Method overview}
We illustrate the pipeline of our method in Fig.~\ref{fig:pipeline}. Our pipeline is composed of a \textbf{L}ocal \textbf{A}ggregation \textbf{F}low \textbf{C}ompletion network (LAFC) for flow completion and an improved version of the flow-guided transformer to synthesize the corrupted regions. Given a corrupted video sequence $X$, we first estimate its bidirectional optical flows $\Tilde{F}_f$ and $\Tilde{F}_b$. Then, we complete each optical flow based on itself and its local references with LAFC. We adopt FGT++ to synthesize the corrupted regions under the guidance of the completed flows. The inference strategy is flexible. We can perform the flow-guide content propagation (FGCP) \cite{Gao-ECCV-FGVC} first and inpaint the rest corrupted regions or synthesize all the corrupted regions with the flow-guided transformer. The former is slower but has better performance. FGT++ inpaints video frames under the sliding window strategy. In each forward pass, we sample local frames $X_l=\{X_{t-s}, ..., X_{t-1}, X_{t}, X_{t+1}, ..., X_{t+s}\}$ and global frames $X_g=\{X_{r}, X_{2r}, ...\}$. Where $s$ denotes the sampling stride of the local frames and $r$ is the sampling interval in global frames. The global frame sequence $X_g$ is used to enlarge the temporal receptive field. We feed $X_l$ and $X_g$ to FGT++ and obtain the completed local frames $\hat{X_l}$. Such a process is iterated until all of the video frames are completed.

\subsection{Flow Completion Network} \label{p3d-unet}
\subsubsection{Local aggregation} The motion and velocity variance caused by force over time causes the degradation of correlation between distant optical flows. Fortunately, the variance of instantaneous motion is a gradual process, therefore the optical flows are highly correlated in a short temporal window. Such correlation can serve as a strong reference for more accurate flow completion.

Previous works \cite{kalluri2023flavr,hara2017learning} mainly adopt 3D convolution \cite{tran2015learning} to capture local temporal correlation. However, it greatly increases the difficulty of network optimization due to its considerable parameter size and computation overhead. Recently, there have been numerous variants of 3D convolutions \cite{ying2020deformable,gu2020appearance,yang2019asymmetric,qiu2017learning}, which maintain the local temporal modeling property of traditional 3D convolution while greatly reducing the computation cost. Considering efficiency, we adopt P3D block \cite{qiu2017learning} instead to capture the local temporal correlation between optical flows in a short temporal window by decoupling the spatial and temporal processing. We integrate P3D blocks into the encoder of LAFC and add skip connection \cite{ronneberger2015u} to exploit the local correlation between optical flows. LAFC processes forward and backward optical flows in the same manner, therefore we unify the signs of forward and backward flows to $F$ for simplicity. Given a corrupted optical flow sequence, we adopt Laplacian filling to obtain the initialized flows $\Tilde{F}$ = \{$\Tilde{F}_{t-ni}, ..., \Tilde{F}_{t}, ..., \Tilde{F}_{t+ni}$\}, where $\Tilde{F}_{t}$ is the target corrupted flow, $i$ is the temporal interval between consecutive flows, and the length of the flow sequence is $2n+1$. We feed the initialized flow sequence $\Tilde{F}$ to LAFC for flow completion of the target flow $\Tilde{F}_{t}$. We denote the input of $m$-th P3D block as $\Tilde{f}^m$, and the output as $\Tilde{f}^{m+1}$. We formulate the local feature aggregation process as
\begin{equation}
    \Tilde{f}^{m+1} = \mbox{TC}(\mbox{SC}(\Tilde{f}^{m})) + \Tilde{f}^{m},
    \label{inertia_quant}
\end{equation} 
where $\mbox{TC}$ represents 1D temporal convolution, and $\mbox{SC}$ is the 2D spatial convolution. We keep the temporal resolution unchanged except for the final P3D block in the encoder and that inserted in the skip connection. We shrink the temporal resolution inside these blocks to obtain the aggregated flow features of the target flow. Finally, we utilize a 2D decoder to obtain the completed target optical flow $\hat{F}_t$.

\subsubsection{Edge loss} \label{EG} As discussed previously, motion variance in a short temporal window is gradual. Therefore, flow fields are piece-wise smooth, which means the gradients of optical flows are considerably small except motion boundaries \cite{Gao-ECCV-FGVC}. The edges in flow maps inherently contain crucial salient features that benefit the reconstruction of motion boundaries. Nevertheless, the reconstruction of motion boundaries in optical flows is tough, as there is no explicit guidance. In LAFC, we design a novel edge loss to supervise the completion quality in motion boundaries of $\hat{F}_t$ explicitly, which can improve the flow completion quality without introducing additional computation overhead during inference.

Firstly, we extract the motion boundaries of the completed target flow $\hat{F}_t$ with a small projection network $P_{e}$. Then, we extract the edges from the ground truth $F_t$ with Canny edge detector \cite{4767851} as the ground truth of motion boundaries and calculate the binary cross entropy loss between them. We formulate edge loss as
\begin{equation}
    L_{e}=\mbox{BCE}(\mbox{Canny}(F_{t}), P_{e}(\hat{F}_{t})).
    \label{edge_loss}
\end{equation} 
We utilize four convolution layers with residual connection \cite{he2016deep} to formulate $P_{e}$.

\subsubsection{Entire loss function} We adopt L1 loss to supervise $\hat{F}_{t}$ in the corrupted and the valid regions, respectively. The loss function is
\begin{equation}
\begin{aligned}
    % & L_{hole} = \frac{\left\|M_t\odot(F_t-\hat{F_T})\right\|_1}{\left\|M_t\right\|_1} \\
    & L_{c} = \left\|M_t\odot(F_t-\hat{F_t})\right\|_1/\left\|M_t\right\|_1, \\
    & L_{v} = \left\|(1-M_t)\odot(F_t-\hat{F_t})\right\|_1/\left\|(1-M_t)\right\|_1,
\end{aligned}
\end{equation}
where $\odot$ denotes as Hadamard product. $L_c$ and $L_{v}$ represent the reconstruction loss in the corrupted and the valid regions, respectively.

Considering the piece-wise smoothness property of the optical flows, we impose the first- and second-order smoothness losses to $\hat{F}_{t}$.
\begin{equation}
   L_{s}=\left\|\nabla \hat{F}_{t}\right\|_1  + \left\|\triangle \hat{F}_{t}\right\|_1.
   \label{smooth}
\end{equation}

Moreover, we also warp the corresponding ground truth frames with $\hat{F}_{t}$ to penalize the regions with large warp errors. We formulate the warp loss as
\begin{equation}
   L_{w}=\left\|\mathcal{W}(\hat{F}_{t\rightarrow t+1}, Y_{t+1}) - Y_{t}\right\|_1.
   \label{warp}
\end{equation}
The warp loss with backward flows can be extended straightforwardly. $\mathcal{W}$ stands for the optical flow-based frame warping with forward-backward consistency check to expel occlusion regions for more robust warping error estimation \cite{lai2018learning}. Specifically, $\mathcal{W}(\hat{F}_{t\rightarrow t+1}, Y_{t+1})$ is to warp $Y_{t+1}$ with the completed optical flow $\hat{F}_{t\rightarrow t+1}$ towards the $t$-th timestamp. For the forward-backward consistency check, given the ground-truth optical flows $F_{t\rightarrow t+1}$ and $F_{t+1\rightarrow t}$, we have 
\begin{equation}
\begin{aligned}
    & k = F_{t\rightarrow t+1}(p), \\
    & p_{end} = k + F_{t+1\rightarrow t}(p+k).
    \label{consist1}
\end{aligned}
\end{equation}
where $p$ is a sampled point, and $p_{end}$ is the point after forward and backward warping. If the Euclidean distance between $p_{end}$ and $p$ is larger than 5, we omit this point in the flow warp loss calculation. This setting follows FGVC~\cite{Gao-ECCV-FGVC} and VINet~\cite{kim2019deep}. 

LAFC adopts the combination of $L_c$, $L_v$, $L_s$, $L_w$ and the edge loss $L_e$ as the loss function, which is formulated as
\begin{equation}
    L_{F}=\lambda_1 L_{c} + \lambda_2 L_{v} + \lambda_3 L_{s} + \lambda_4 L_{w} + \lambda_5 L_{e}.
    \label{loss_f}
\end{equation}
We simply set $\lambda_1$, $\lambda_2$ and $\lambda_5$ to 1, $\lambda_3$ to 0.5 and $\lambda_4$ to 0.01 for the balance of magnitude between different loss terms.

\subsection{Overview of Flow-Guided Transformer} The framework of improved flow-guided transformer is illustrated in Fig.~\ref{fig:pipeline}. The inputs are the combination of local and global frames $X_{in}=X_l \cup X_g$. These frames are firstly encoded to feature space $FI$, and then transformed to tokens maps $TI$ for transformer processing. The kernel size, stride, and padding in the transformation operation follow the settings of FFM~\cite{Liu_2021_FuseFormer}. We follow CVPT \cite{chu2021conditional} to provide positional embedding for video inpainting in flexible
resolutions, which adopts a depth-wise
convolution block \cite{howard2017mobilenets} after the first transformer block. We utilize the spatiotemporally decoupled transformer blocks to explore the feature correlation to complete the corrupted features. Finally, we utilize a decoder to output the completed frames $\hat{Y}$. 

In order to address the query degradation problem, we utilize the bidirectional completed optical flows of $X_l$ to build the correlation between complementary features in $FI$ for high-quality feature generation in the corrupted regions. According to the intrinsic properties of spatial and temporal processing, we elaborately design different window partition strategies for temporal and spatial transformer blocks. In temporal transformer blocks, we combine large window and temporal deformable MHSA for attention retrieval in different temporal granularity. In spatial transformer blocks, we design the dual perspective MHSA to maintain the local smoothness while enlarging the spatial receptive field to strike a balance between performance and efficiency tradeoff.

\begin{figure}[t]
\begin{center}
\includegraphics[width=0.9\linewidth]{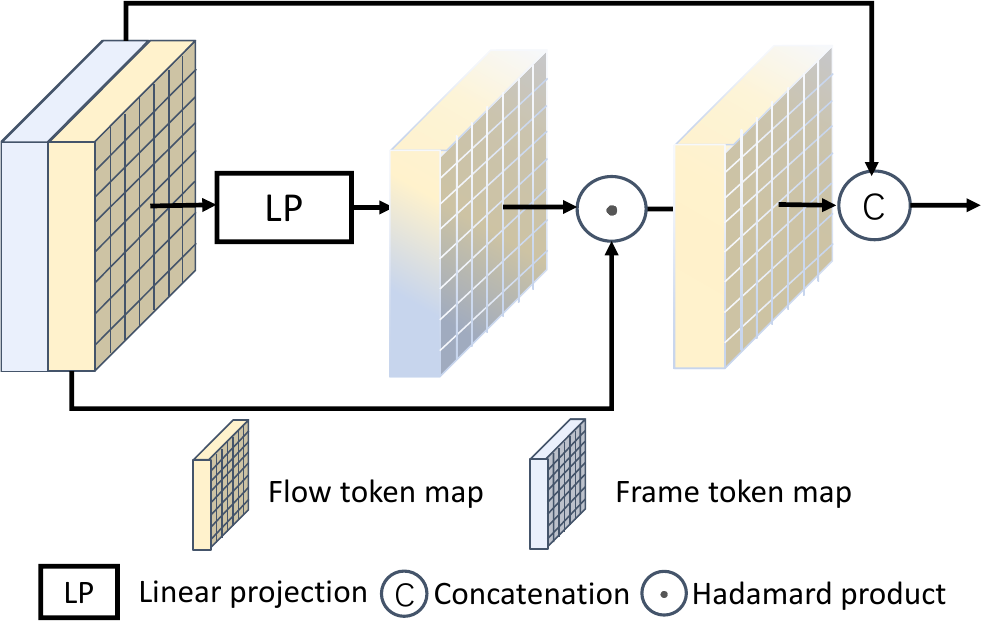}
\end{center}
   \caption{Illustration of the proposed flow guidance feature integration (FGFI) module.}
\label{fig:fgfi}
\end{figure}

\subsection{Proposed Solutions for Query Degradation}
\subsubsection{Query degradation} Previous transformer-based video inpainting methods \cite{yan2020sttn,liu2021decoupled,Liu_2021_FuseFormer} mainly contain a frame-wise encoder, some cascaded transformer blocks, and a frame-wise decoder. The MHSA mechanism in transformer blocks measures the cosine similarity between the query and key tokens, which is formulated as
\begin{equation}
    \boldsymbol{H}={\rm softmax}(\boldsymbol{Q}\boldsymbol{K}^T / \sqrt{D})\boldsymbol{V},
\end{equation}
where $\boldsymbol{H}$, $\boldsymbol{Q}$, $\boldsymbol{K}$ and $\boldsymbol{V}$ represent the completed features after attention processing, query, key, and value, respectively. And $D$ is the token dimension. $\boldsymbol{A}={\rm softmax}(\boldsymbol{Q}\boldsymbol{K}^T / \sqrt{D})$ is the attention score. The query tokens tend to retrieve the key tokens with similar context. If the query token $\boldsymbol{Q}_i$ is closer to the key token $\boldsymbol{K}_j$ than $\boldsymbol{K}_m$ in token space, the corresponding attention score will be $A_{ij} > A_{im}$, where $A_{ij}={\rm softmax}(\boldsymbol{Q}_i\boldsymbol{K}_j^T / \sqrt{D}$).

If the features that generate query tokens are encoded from a frame-wise encoder, the features in the corrupted regions are synthesized based on the features from the valid regions in the same frame. Such operation limits the feature quality in the corrupted regions, especially when the appearance in the corrupted features varies a lot against that in the valid features.

We name such phenomenon as \textit{query degradation} and propose two orthogonal strategies to mitigate this problem. One is ``Flow guidance feature integration" (FGFI) and the other is ``Flow-guided feature propagation" (FGFP).

\subsubsection{Flow guidance feature integration} The motion discrepancy between different objects and the background exposed in completed optical flows supplies content relationship within the feature map. The tokens with similar motion magnitude are more likely to be relevant. Thus the completed flows are capable of serving as the additional guidance to enhance the frame features for more accurate attention retrieval.

As discussed in Sec.~\ref{p3d-unet}, the local correlated property of different optical flows lacks temporally long-range modeling ability, which is necessary for temporal attention retrieval. Therefore, we decouple MHSA in transformer blocks along spatial and temporal dimensions and only exploit the optical flows to enhance the features in the spatial transformer to perform flow guidance feature integration. A straightforward way is to encode completed flows $\hat{F}_t$ into flow tokens $TF$ and concatenate with the frame tokens $TI$ along channel dimension before performing spatial MHSA. However, the imperfectness of completed flows may mislead the judgment of the relevant regions. Moreover, similar motion patterns may indicate different appearances and the appearance may also vary a lot within objects, which is likely to confuse the attention retrieval process. Therefore, we propose a flow-reweight module to control the impact of flow tokens $TF$ based on the interaction between $TF$ and $TI$. We illustrate the flow token integration process in Fig.~\ref{fig:fgfi} and formulate the process as
\begin{equation}
    \hat{TF}_t^j = TF_t^j \odot \mbox{MLP}(\mathcal{C}(TI_t^j, TF_t^j)),
\end{equation}
where $\mathcal{C}$ is the concatenation operation. $\mbox{MLP}$ stands for the MLP layers. $TI_t^j$, $TF_t^j$ and $\hat{TF}_t^j$ represent the $t$-th frame token map, flow token map, and reweighted flow token map in $j$-th spatial transformer, respectively. Finally, we concatenate $\hat{TF}_t^j$ and $TI_t^j$ to obtain the flow-enhanced tokens $\Tilde{TI}_t^j$ to guide attention retrieval in spatial MHSA.

\begin{figure}[t]
\begin{center}
\includegraphics[width=1\linewidth]{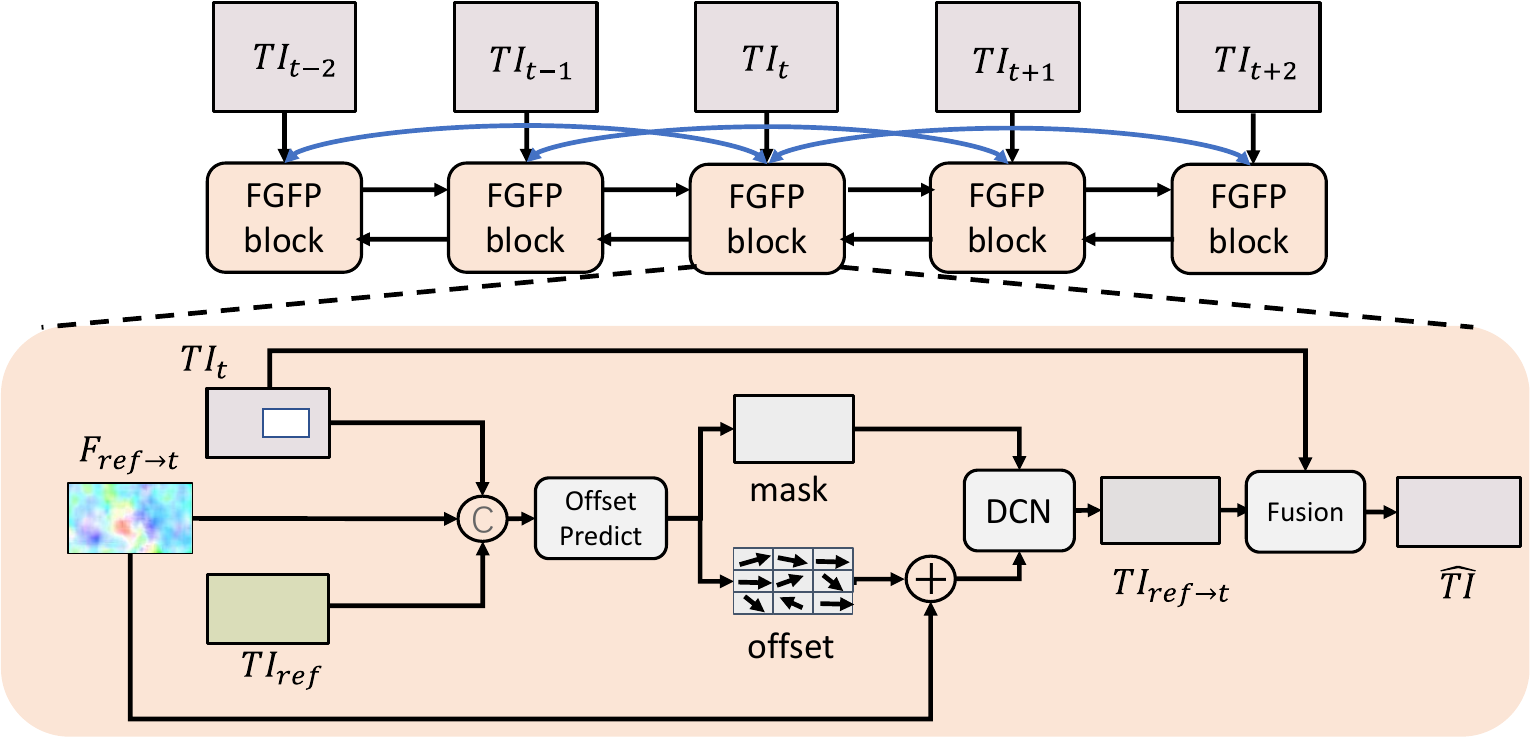}
\end{center}
   \caption{Procedure of the proposed flow-guided feature propagation (FGFP) module. All the local frames share the same FGFP block. We illustrate the details of one FGFP block between a single reference frame and the target frame for simplicity.}
\label{fig:fgfp}
\vspace{-0.5cm}
\end{figure}

\subsubsection{Flow-guided feature propagation} The relative motion between the corrupted regions and the scenes causes the exposure of complementary features across different frames. Therefore, modeling the correlation of features temporally and propagating them along the temporal dimension is crucial for obtaining accurate features. Based on such motivation, we design the flow-guided feature propagation module (FGFP).

Fig.~\ref{fig:fgfp} depicts the working procedure of FGFP and the technical details for feature propagation. Inspired by \cite{chan2022basicvsrpp,liCvpr22vInpainting}, we adopt the combination of first-order, second-order, and bidirectional feature propagation for more robust propagation performance. We observed that higher-order feature propagation provides a marginal improvement in the FGFP module (the related experimental results are provided in the supplementary material). We only propagate the features across the features from local frames $X_l$ because the quality of optical flows across temporally distant frames degrades severely.  Therefore, we perform content propagation across only the local frames $X_l$. The global frames $X_g$ are not involved in the FGFP module. Assume the $t$-th feature in $FI$ is $FI_t$, the completed optical flow from $t$ to $t+1$ is $\hat{F}_{t\rightarrow t+1}$, the FGFP module can be formulated as
\begin{equation}
\begin{aligned}
    \hat{FI}_t^f &= \mbox{FGFP}(FI_t, FT_{t-1}, \hat{F}_{t\rightarrow t-1}, FT_{t-2}, \hat{F}_{t\rightarrow t-2}), \\
    \hat{FI}_t^b &= \mbox{FGFP}(FI_t, FT_{t+1}, \hat{F}_{t\rightarrow t+1}, FT_{t+2}, \hat{F}_{t\rightarrow t+2}), \\
    \hat{FI}_t &= \mathcal{E}(\hat{FT}_t^f, \hat{FT}_t^b),
\end{aligned}
\end{equation}
where $\hat{FI}_t^f$ and $\hat{FI}_t^b$ denote the forward and backward propagated features, respectively, and $\hat{FI}_t$ is the propagated feature after feature fusion. $\mbox{FGFP}$ is the FGFP block. All the local frames in $X_l$ share the same FGFP block. And $\mathcal{E}$ is the feature fusion module to aggregate the propagated features.

In order to compensate for the distorted motion trajectory from the imperfect completed optical flows, we adopt the deformable convolution \cite{zhu2019deformable} to predict the residual motion offset of the completed optical flows. The trainable deformable convolution can refine the distortion of completed optical flows. Such refinement operation is critical for the feature propagation accuracy, which leads to further video inpainting performance improvement.

\begin{figure}[t]
\begin{center}
\includegraphics[width=0.9\linewidth]{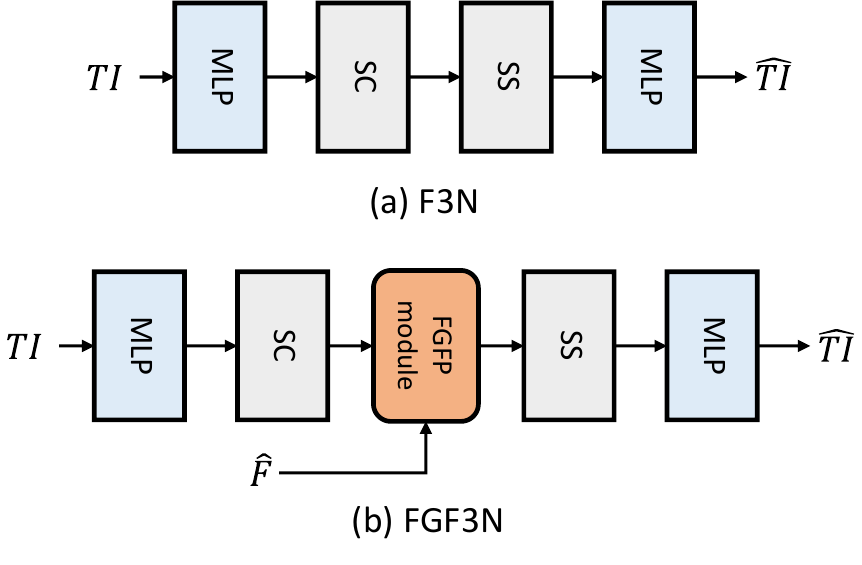}
\end{center}
   \caption{Comparison between F3N and FGF3N modules. SC and SS stand for the soft composition and soft split operations in FFM~\cite{Liu_2021_FuseFormer}, respectively. $\hat{F}$ denotes the completed optical flows.}
\label{fig:fgf3n}
\end{figure}

We place the flow-guided feature propagation (FGFP) module before MHSA in transformer blocks. For the first transformer block, we insert the FGFP module between the frame-wise encoder and the transformer block. For other transformer blocks, if we directly insert FGFP between two transformer blocks, the extra transition between feature space and token space will bring additional computational costs. The feed-forward layer in FGT~\cite{zhang2022flow} follows the F3N layer in FFM~\cite{Liu_2021_FuseFormer}, which contains the transition process between these two spaces. The formula of F3N is
\begin{equation}
\begin{aligned}
    TI_t^1 &= \mbox{MLP}(TI_t), &FI_t^1 &= \mbox{SC}(TI_t^1), \\
    \hat{TI}_t^1 &= \mbox{SS}(FI_t^1), &\hat{TI}_t &= \mbox{MLP}(\hat{TI}_t^1), \\
\end{aligned}
\end{equation}
where $TI_t^1$ is token map processed by the first $\mbox{MLP}$ layer; $FI_t^1$ is generated from $TI_t^1$ with soft composition ($\mbox{SC}$); $\hat{TI}_t^1$ is the token map processed by soft split ($\mbox{SS}$) and $\hat{TI}_t$ is the output feature from the second $\mbox{MLP}$ layer. The transition between token space and feature space in F3N enables us to integrate FGFP modules into the transformer blocks. We place the FGFP module between SC and SS in F3N, as depicted in Fig.~\ref{fig:fgf3n}. We denote the modified F3N layer as the FGF3N layer (flow-guided F3N layer). The FGFP module in the previous transformer is capable of improving the feature expressiveness, which is beneficial to ameliorating the attention retrieval accuracy in the MHSA of the later transformer blocks.

\begin{figure}[t]
\begin{center}
\includegraphics[width=0.7\linewidth]{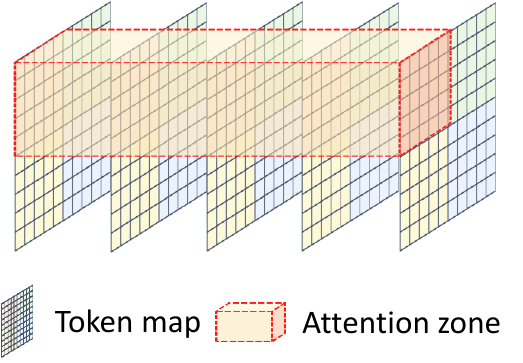}
\end{center}
   \caption{Illustration of the large-window MHSA in the temporal transformer blocks.}
\label{fig:lw}
\end{figure}

\subsection{Flow-Guided Transformer Architecture}
\subsubsection{Temporally deformable MHSA} FGT \cite{zhang2022flow} designs a novel large window transformer to maintain a large spatiotemporal receptive field while reducing the computation overhead compared with all-pair attention retrieval in FFM~\cite{Liu_2021_FuseFormer}. The core of this transformer block is large window MHSA, as depicted in Fig.~\ref{fig:lw}, which plays an important role in aggregating the context features from temporally distant frames. 

However, large window MHSA is inefficient in modeling the feature correlation between the temporally nearby frames. The motion fields in temporally nearby frames are powerful priors in capturing temporal feature correspondence, which is naturally described by the completed optical flows $\hat{F}$. Therefore, besides the temporally global large window MHSA, we design a novel temporal deformable MHSA (TD-MHSA) to refine the attention retrieval in local frames $X_l$ under the guidance from completed flows $\hat{F}$. Since TD-MHSA aims at capturing the correspondences of the correlated tokens under the guidance of the completed optical flows, only local frames $X_l$ are involved in the process of TD-MHSA to avoid the inaccuracy of motion fields due to distant frames.

\begin{figure}[t]
\begin{center}
\includegraphics[width=0.9\linewidth]{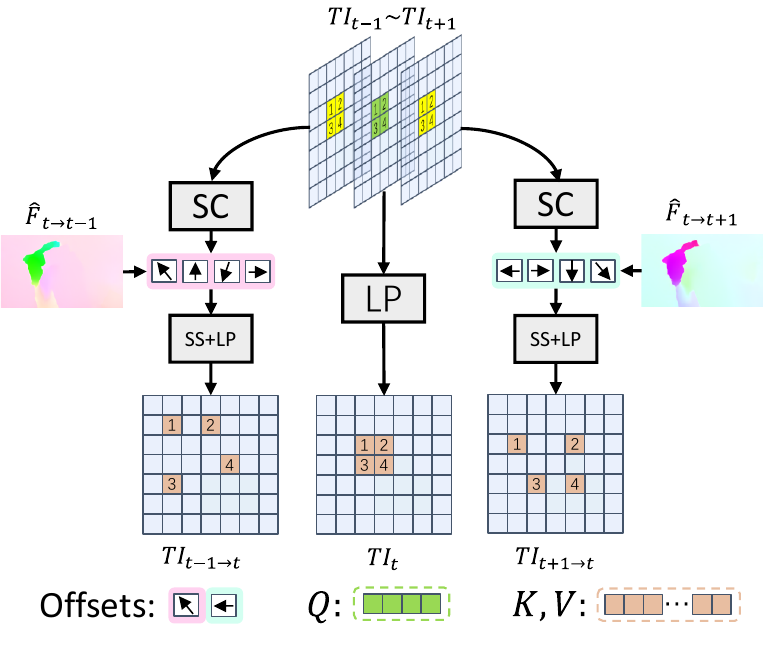}
\end{center}
   \caption{Details of the temporal deformable MHSA in the temporal transformer blocks. SS: Soft split; SC: Soft composition; LP: Linear projection.}
\label{fig:td}
\end{figure}

We illustrate the module design of TD-MHSA in Fig.~\ref{fig:td}. TD-MHSA aims to model the correlation between the $t$-th token map $TI_t$ and the relevant content in $t-1$-th and $t+1$-th token maps guided by completed optical flows $\hat{F}_{t\rightarrow t-1}$ and $\hat{F}_{t \rightarrow t+1}$. First, we use SC to transform the token map $TI_{t-1}$ and $TI_{t+1}$ to the feature space and obtain the corresponding features
\begin{equation}
\begin{aligned}
    \hat{FI}_{t-1} &= \mbox{SC}(TI_{t-1}), \\
    \hat{FI}_{t+1} &= \mbox{SC}(TI_{t+1}).
\end{aligned}
\end{equation}

After we obtain $\hat{FI}_{t-1}$ and $\hat{FI}_{t+1}$, we exploit the corresponding completed optical flows $\hat{F}_{t\rightarrow t-1}$ and $\hat{F}_{t \rightarrow t+1}$ to aggregate the relevant content in the warped location along the temporal dimension, and use SS to map the features to token space for attention.
\begin{equation}
\begin{aligned}
    TI_{t-1 \rightarrow t} &= \mbox{SS}(\mathcal{W}(\hat{FI}_{t-1}, \hat{F}_{t \rightarrow t-1})), \\
    TI_{t+1 \rightarrow t} &= \mbox{SS}(\mathcal{W}(\hat{FI}_{t+1}, \hat{F}_{t \rightarrow t+1})),
\end{aligned}
\end{equation}
where $\mathcal{W}$ is the backward warping operation, $TI_{t-1 \rightarrow t}$ and $TI_{t+1 \rightarrow t}$ represent the warped token map from $t-1$ to $t$ and $t+1$ to $t$, respectively.

Finally, we encode the query, key, and value based on the warped token maps. An ideal case is that we only perform MHSA for the tokens in the corresponding spatial location along the temporal dimension. However, since the completed optical flows are not perfect, the constructed feature relevance may be inaccurate. To improve the robustness of TD-MHSA, we impose the window partition strategy on the token maps. Different from large window MHSA, we adopt smaller windows to TD-MHSA because the overall correspondence has been modeled by the completed optical flows. We set the height and width of TD-MHSA to be half of that in large window MHSA. For $t$-th token map $TI_t$ encoded from local frames $X_l$, we denote its $d$-th window as $TI_t(d)$. We formulate the encoding process of query, key, and value as
\begin{equation}
\begin{aligned}
    Q_t(d) &= \mbox{MLP}(\mbox{LN}(TI_t(d)), \\
    K_t(d) &= \mbox{MLP}(\mbox{LN}(\mathcal{C}(\hat{TI}_{t-1 \rightarrow t}(d), TI_t(d), \hat{TI}_{t+1 \rightarrow t}(d)))), \\
    V_t(d) &= \mbox{MLP}(\mbox{LN}(\mathcal{C}(\hat{TI}_{t-1 \rightarrow t}(d), TI_t(d), \hat{TI}_{t+1 \rightarrow t}(d)))),
\end{aligned}
\end{equation}
where $\mbox{MLP}$ and $\mbox{LN}$ represents the MLP and layer normalization layers \cite{ba2016layer}, respectively. $\mathcal{C}$ is the concatenation operation. $Q_t(d)$, $K_t(d)$ and $V_t(d)$ denotes as the query, key and value tokens in $d$-th window, respectively. We perform MHSA based on the encoded query, key, and value.

\subsubsection{Dual perspective MHSA}
Since the neighbor tokens are more correlated due to the local smoothness prior of natural images, we adopt a relatively small window in the spatial transformer. However, a small window size in the spatial transformer leads to a limited receptive field, which undermines the quality of attention retrieval of the spatial transformer. To mitigate this problem, we design dual perspective MHSA, where we input the token maps from both the local and the global frames. Here, we use $TI$ to denote the token maps from both local and global frames. Assume the height, width, and channel size of the token map are $H$, $W$ and $C$, respectively, the $t$-th frame token can be formulated as $TI_t^j \in \mathbb{R}^{H \times W \times C}$. We divide $TI_t^j$ into several $h \times w$ non-overlapped windows and perform spatial MHSA inside each window. However, such an operation introduces a restricted receptive field, which causes sub-optimal attention retrieval performance. Moreover, if the window contains numerous tokens projected from the corrupted regions, such a restricted receptive field lacks the ability to retrieve the content from the valid regions. Therefore, we adopt depth-wise convolution \cite{howard2017mobilenets} to condense $TI_t^j$ to global tokens, and feed them to each window. Given the kernel size $k$ and downsampling rate $s$ (also known as stride), the global tokens are generated as
\begin{equation}
    TG_t=\mbox{DC}(TI_t, k, s),
    \label{TG_gen}
\end{equation}
where $TG_t$ represents the condensed global tokens and $\mbox{DC}$ is the depth-wise convolution. The query $Q_t(d)$, key $K_t(d)$ and value $V_t(d)$ of the $d$-th window in $TI_t$ are generated as
\begin{figure}[t]
\begin{center}
\includegraphics[width=0.9\linewidth]{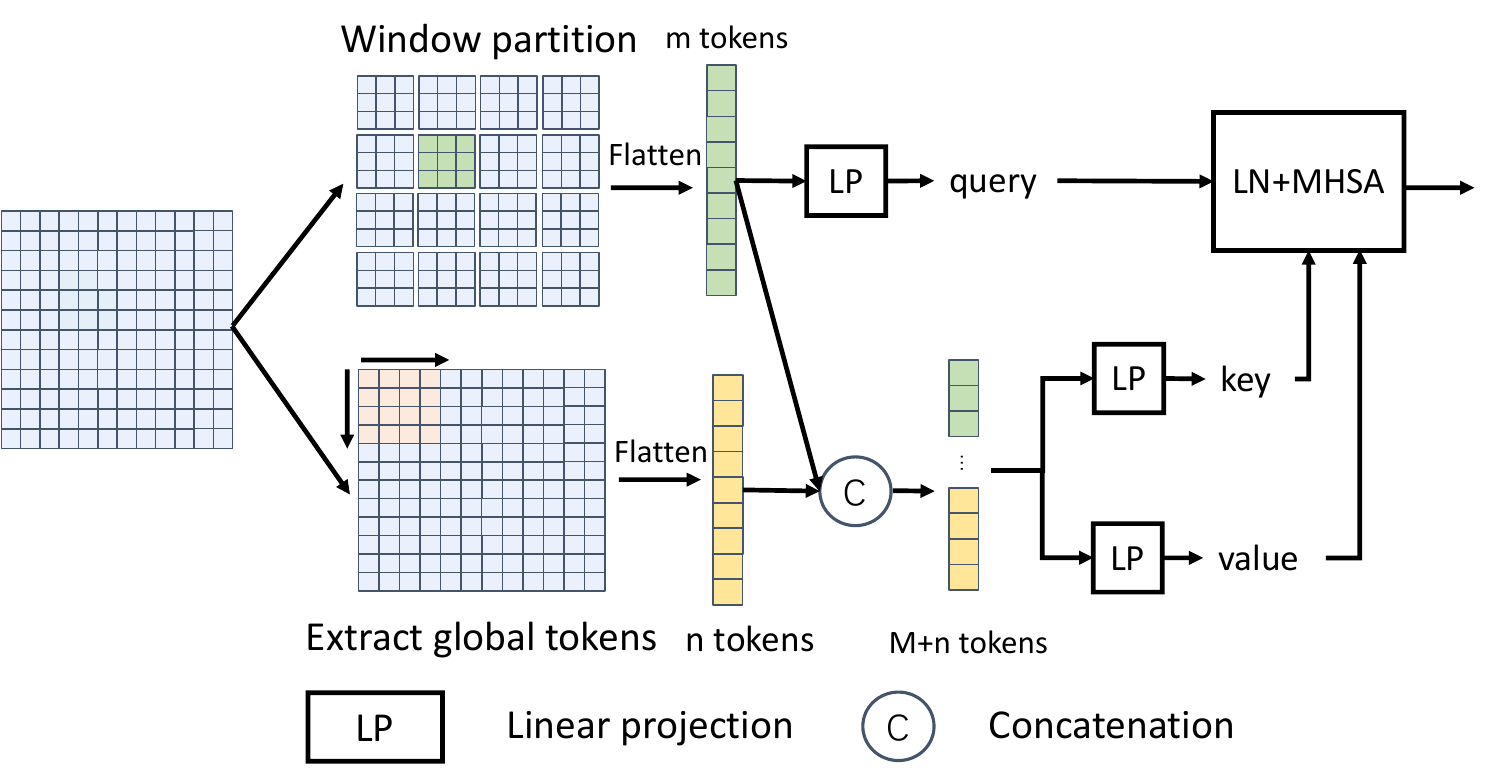}
\end{center}
   \caption{Illustration of the dual perspective spatial MHSA in the spatial transformer blocks.}
\label{fig:dpmhsa}
\end{figure}

\begin{equation}
\begin{aligned}
    & Q_t(d) = \mbox{MLP}(\mbox{LN}(TI_t(d))), \\
    & K_t(d) = \mbox{MLP}(\mbox{LN}(\mathcal{C}(TI_t(d),TG_t))), \\
    & V_t(d) = \mbox{MLP}(\mbox{LN}(\mathcal{C}(TI_t(d),TG_t))),
\end{aligned}
\end{equation}
where $TI_t(d)$ stands for the $d$-th window in $TI_t$. We apply the spatial MHSA to process $Q_t(d)$, $K_t(d)$ and $V_t(d)$. We illustrate the dual perspective spatial MHSA in Fig.~\ref{fig:dpmhsa}.

Note that the global tokens are shared by all the windows. In each spatial transformer, all-pair attention retrieval in MHSA will lead each token to retrieve $H \times W$ tokens. While the token number for retrieval in our dual perspective spatial MHSA is $(\lceil\frac{H}{s}\rceil \times \lceil\frac{W}{s}\rceil + h \times w)$. We can derive that when $s > \lceil\sqrt{\frac{HW}{HW-hw}}\rceil$, the referenced token number will get reduced compared with all-pair attention retrieval.

Recently, focal transformer \cite{yang2021focal} also combines local and global attention in the transformer. Compared with \cite{yang2021focal}, our method decouples the global token size and the window shape, which is more flexible than the sub-window pooling strategy in the focal transformer.

\subsubsection{Loss function} We adopt the reconstruction loss in the corrupted and the valid regions together with the T-Patch GAN loss \cite{chang2019free} to supervise the training process. Different from previous works \cite{yan2020sttn,Liu_2021_FuseFormer,liCvpr22vInpainting}, we measure the reconstruction loss not only from the spatial domain, but we also introduce the frequency domain loss to supervise the amplitude of Fourier transform between the reconstructed frames and the ground truth. The spatial domain reconstruction loss can be formulated as
\begin{equation}
\begin{aligned}
    & L_{yc} = \left\|M_t\odot(Y_t-\hat{Y_t})\right\|_1/\left\|M_t\right\|_1, \\
    & L_{yv} = \left\|(1-M_t)\odot(Y_t-\hat{Y_t})\right\|_1/\left\|(1-M_t)\right\|_1.
\end{aligned}
\end{equation}

Recently, the analysis in the frequency domain is introduced in the low-level vision field, and achieved significant progress in image inpainting \cite{suvorov2022resolution}, video super-resolution \cite{qiu2022learning}, image enhancement \cite{huang2022deep}, etc. To the best of our knowledge, this work is the first attempt to introduce frequency domain analysis in video inpainting. 

Specifically, we introduce amplitude loss to the training of FGT++, which supervises the amplitude of the inpainted frame $\hat{Y}_t$ and the ground truth $Y_t$. For a given image $K$, the Fourier transform converts $K$ from the spatial to the frequency domain.
\begin{equation}
    K(u,v) = \frac{1}{\sqrt{HW}}\sum_{h=0}^{H-1}\sum_{w=0}^{W-1}K(h,w)e^{-j2\pi (\frac{h}{H}u+\frac{w}{W}v)}.
\end{equation}

We denote the above process as $\mathcal{F}$. After we obtain $K(u,v)$, we solve the amplitude component with the real and imaginary part of $K(u,v)$
\begin{equation}
    \mathcal{A}(K(u,v))=\sqrt{R^2(K(u,v))+I^2(K(u,v))},
\end{equation}
where $R(K(u,v))$ and $I(K(u,v))$ represents the real and imaginary part of $K(u,v)$, respectively. And $\mathcal{A}(K(u,v))$ denotes as the amplitude of $K(u,v)$. Following the above process, we can get the amplitude of the restored frame and the ground truth, and supervise the distance between them with $L_1$ loss, which is formulated as
\begin{equation}
    L_{amp}=\left\|\mathcal{A}(\mathcal{F}(\hat{Y}_t)) - \mathcal{A}(\mathcal{F}(Y_t))\right\|_1.
\end{equation}

As for adversarial loss, we follow the previous work \cite{yan2020sttn}, which is formulated as
\begin{equation}
    L_{adv} = -\mathbb{E}_{z \sim P_{\hat{Y}_{t}(z)}}[D(z)].
\end{equation}
The discriminator loss is
\begin{equation}
\begin{aligned}
    L_D &= \mathbb{E}_{x \sim P_{Y_{t}}(x)}[\mbox{ReLU}(1+D(x))], \\ 
    &+ \mathbb{E}_{z \sim P_{\hat{Y}_{t}}(z)}[\mbox{ReLU}(1-D(z))]. 
\end{aligned}
\end{equation}
where $D$ represents the discriminator. Therefore, the generator loss is the combination of the loss terms described above.
\begin{equation}
    L_{y}=\lambda_{y1} L_{yc} + \lambda_{y2} L_{yv} + \lambda_{y3} L_{adv} + \lambda_{y_4} L_{amp}.
    \label{loss_y}
\end{equation}
Following previous works \cite{yan2020sttn,Liu_2021_FuseFormer}, we simply set $\lambda_{y1}$ and $\lambda_{y2}$ to 1, $\lambda_{y3}$ to 0.01 and $\lambda_{y_4}$ to 0.1. 
% \begin{figure*}[t]
% \begin{center}
% \includegraphics[width=0.6\linewidth]{figures/temporalTrans_jing2.pdf}
% \end{center}
%    \caption{The temporal MHSA in the temporal transformer. We split non-overlapped large windows (zones) for each token map, and perform MHSA inside the cube formed by the corresponding position in each token map. The windows are shown with different colors. In this figure, we illustrate the 2$\times$2 zone.}
% \label{fig:temporal}
% \end{figure*}

% \begin{figure*}[t]
% \begin{center}
% \includegraphics[width=1\linewidth]{figures/spatialTrans_jing3.pdf}
% \end{center}
%    \caption{Illustration of flow guidance integration and dual perspective spatial MHSA in the spatial transformer.}
% \label{fig:spatial}
% \end{figure*}

\begin{table*}[t]
\fontsize{10}{10}\selectfont
\begin{center}
% \scriptsize
% \small
\caption{Quantitative results on the Youtube-VOS and DAVIS datasets. The best and second best numbers for each metric are indicated by \textcolor[rgb]{1.00,0.00,0.00}{\underline{red}} and \textcolor[rgb]{0.00,0.00,1.00}{\underline{blue}} respectively. $\downarrow$ means lower is better, while $\uparrow$ means higher is better. ``FGT++*" means that we adopt the completed flows to perform the flow-guided content propagation and then utilize FGT++ to fill the rest unfilled regions.}
\label{quan_rets}
\begin{tabular}{@{}lccccccccc@{}}
\toprule
\multirow{3}{*}{Method} & \multicolumn{3}{c}{\multirow{2}{*}{Youtube-VOS}} & \multicolumn{6}{c}{DAVIS} \\ \cline{5-10}
& \multicolumn{3}{c}{} & \multicolumn{3}{c}{square} & \multicolumn{3}{c}{object} \\ \cmidrule(l){2-4} \cmidrule(l){5-7} \cmidrule(l){8-10}
& PSNR$\uparrow$ & SSIM$\uparrow$ & LPIPS$\downarrow$ & PSNR$\uparrow$ & SSIM$\uparrow$ & LPIPS$\downarrow$ & PSNR$\uparrow$ & SSIM$\uparrow$ & LPIPS$\downarrow$ \\ \midrule
VINet \cite{kim2019deep}& 29.83 & 0.955 & 0.047 & 28.32 & 0.943 & 0.049 & 28.47 & 0.922 & 0.083 \\
DFGVI \cite{Xu_2019_CVPR} &32.05 & 0.965 & 0.038 & 29.75 & 0.959 & 0.037 & 30.28 & 0.925 & 0.052 \\
CPN \cite{lee2019cpnet} &32.17 & 0.963 & 0.040 & 30.20 & 0.953 & 0.049 & 31.59 & 0.933 & 0.058 \\
OPN \cite{Oh_2019_ICCV} &32.66 & 0.965 & 0.039 & 31.15 & 0.958 & 0.044 & 32.40 & 0.944 & 0.041 \\
3DGC \cite{chang2019free} &30.22 & 0.961 & 0.041 & 28.19 & 0.944 & 0.049 & 31.69 & 0.940 & 0.054 \\
STTN \cite{yan2020sttn} &32.49 & 0.964 & 0.040 & 30.54 & 0.954 & 0.047 & 32.83 & 0.943 & 0.052  \\
TSAM \cite{zou2020progressive}& 31.62 & 0.962 & 0.031 & 29.73 & 0.951 & 0.036 & 31.50 & 0.934 & 0.048 \\
DSTT \cite{liu2021decoupled}& 33.53 & 0.969 & 0.031 & 31.61 & 0.960 & 0.037 & 33.39 & 0.945 & 0.050 \\
FFM \cite{Liu_2021_FuseFormer}& 33.73 & 0.970 & 0.030 & 31.87 & 0.965 & 0.034 & 34.19 & 0.951 & 0.045 \\
E2FGVI \cite{liCvpr22vInpainting} & 34.75 & 0.974 & 0.027 & 33.06 & 0.969 & 0.030 & 35.02 & 0.957 & 0.039 \\
FGVC \cite{Gao-ECCV-FGVC} & 33.94 & 0.972 & 0.026 & 32.14 & 0.967 & 0.030 & 33.91 & 0.955 & 0.036 \\
FGT \cite{zhang2022flow} & 34.04 & 0.971 & 0.028 & 32.60 & 0.965 & 0.032 & 34.30 & 0.953 & 0.040 \\
FGT++ & \textcolor[rgb]{0.00,0.00,1.00}{\underline{35.02}} & \textcolor[rgb]{0.00,0.00,1.00}{\underline{0.976}} & \textcolor[rgb]{0.00,0.00,1.00}{\underline{0.025}} & \textcolor[rgb]{0.00,0.00,1.00}{\underline{33.18}} & \textcolor[rgb]{0.00,0.00,1.00}{\underline{0.971}} & \textcolor[rgb]{0.00,0.00,1.00}{\underline{0.028}} & \textcolor[rgb]{0.00,0.00,1.00}{\underline{35.61}} & \textcolor[rgb]{0.00,0.00,1.00}{\underline{0.961}} & \textcolor[rgb]{0.00,0.00,1.00}{\underline{0.035}} \\
FGT++* &\textcolor[rgb]{1.00,0.00,0.00}{\underline{35.36}} & \textcolor[rgb]{1.00,0.00,0.00}{\underline{0.978}} & \textcolor[rgb]{1.00,0.00,0.00}{\underline{0.022}} & \textcolor[rgb]{1.00,0.00,0.00}{\underline{33.72}}& \textcolor[rgb]{1.00,0.00,0.00}{\underline{0.976}} & \textcolor[rgb]{1.00,0.00,0.00}{\underline{0.022}} & \textcolor[rgb]{1.00, 0.00, 0.00}{\underline{35.90}} & \textcolor[rgb]{1.00,0.00,0.00}{\underline{0.968}} & \textcolor[rgb]{1.00,0.00,0.00}{\underline{0.027}}  \\
\bottomrule
\end{tabular}
\end{center}
\end{table*}

\section{Experiments}
\subsection{Settings}
We adopt Youtube-VOS \cite{xu2018youtube} and DAVIS \cite{caelles20182018} datasets for evaluation. Youtube-VOS contains 4453 videos, including 3471 videos for training, 474 videos for validation, and 508 videos for inference. DAVIS contains 150 videos, including 60 for training and 90 for validation. Following the original splits, we adopt the training set of Youtube-VOS to train our networks and the testset for inference. As for DAVIS, we adopt its training set for inference because these frames have corresponding densely annotated masks.

Following the previous work \cite{Gao-ECCV-FGVC}, we choose PSNR, SSIM \cite{wang2004image}, and LPIPS \cite{zhang2018unreasonable} as video inpainting metrics. Meanwhile, we adopt end-point-error (EPE) to evaluate the flow completion quality. We compare our method with FGT \cite{zhang2022flow} and other state-of-the-art baselines, including VINet \cite{kim2019deep}, DFGVI \cite{Xu_2019_CVPR}, CPN \cite{lee2019cpnet}, OPN \cite{Oh_2019_ICCV}, 3DGC \cite{chang2019free}, STTN \cite{yan2020sttn}, FGVC \cite{Gao-ECCV-FGVC}, TSAM \cite{zou2020progressive}, DSTT \cite{liu2021decoupled}, FFM \cite{Liu_2021_FuseFormer} and E2FGVI \cite{liCvpr22vInpainting}.

\subsection{Implementation Details}
In our experiments, We utilize RAFT \cite{teed2020raft} to estimate optical flows. In LAFC, the flow interval and input flow number are both set to 3. The middle optical flow is treated as the target flow during completion. FGT++ adopts 8 transformer blocks in total (4 temporal and 4 spatial transformer blocks). In the temporal transformer blocks, we adopt 2$\times$2 zone division for large window MHSA, and the height and width of temporal deformable MHSA are set to be half of that in large window MHSA. In the spatial transformer blocks, the downsampling rate of the global token is 4, while the window size is 8. Different from FGT, FGT++ adopts the FGFI module only in the first spatial transformer block. We adopt the FGFP module in the first 6 transformer blocks and the TD-MHSA to all the temporal transformer blocks. We adopt Adam optimizer \cite{kingma2014adam} to train our networks. The training iteration is 280K for LAFC and 500K for FGT++. The initial learning rate is 1$e$-4, which is divided by 10 after 120K iterations for LAFC and 400K iterations for FGT++. During FGT++ training, we sample 5 temporally nearby frames as the local frames and sample an additional 3 frames as the global frames. For ablation studies, following FFM \cite{Liu_2021_FuseFormer}, we choose the DAVIS dataset, and train FGT++ for 250K iterations, whose learning rate is divided by 10 after 200K iterations.

\subsection{Quantitative Evaluation}
We set the resolution of videos to 432$\times$256 during inference. In order to evaluate the performance comprehensively, we adopt the square maskset and object maskset during inference. The square maskset is static or generated with continuous motion trace. We adopt it to evaluate the performance of Youtube-VOS and DAVIS datasets. The average size of the masks is $\frac{1}{16}$ of the whole frame. As for the object maskset, we shuffle DAVIS object masks randomly to evaluate video inpainting performance. For fair comparisons among flow-based video inpainting methods, we utilize the same optical flow extractor for DFGVI \cite{Xu_2019_CVPR} and FGVC \cite{Gao-ECCV-FGVC} as our method. 

We report the quantitative evaluation results between our method and other baselines in Tab.~\ref{quan_rets}. Our method outperforms previous baselines by a significant margin on all three metrics, which means our method is capable of inpainting videos with less distortion and better perceptual quality against existing baselines. The quantitative improvement of FGT++ against FGT demonstrates the effectiveness of the newly proposed components. Compared with pure FGT++, if we adopt flow-guided content propagation first and utilize the FGT++ to inpaint the left corrupted regions, we can boost the performance further, as indicated in FGT++*.
\begin{figure*}[t]
\captionsetup{font={scriptsize}}
    \centering
    \subfigure[(a) Input]{\begin{minipage}[t]{0.129\linewidth}
        \centering\includegraphics[width=1\linewidth]{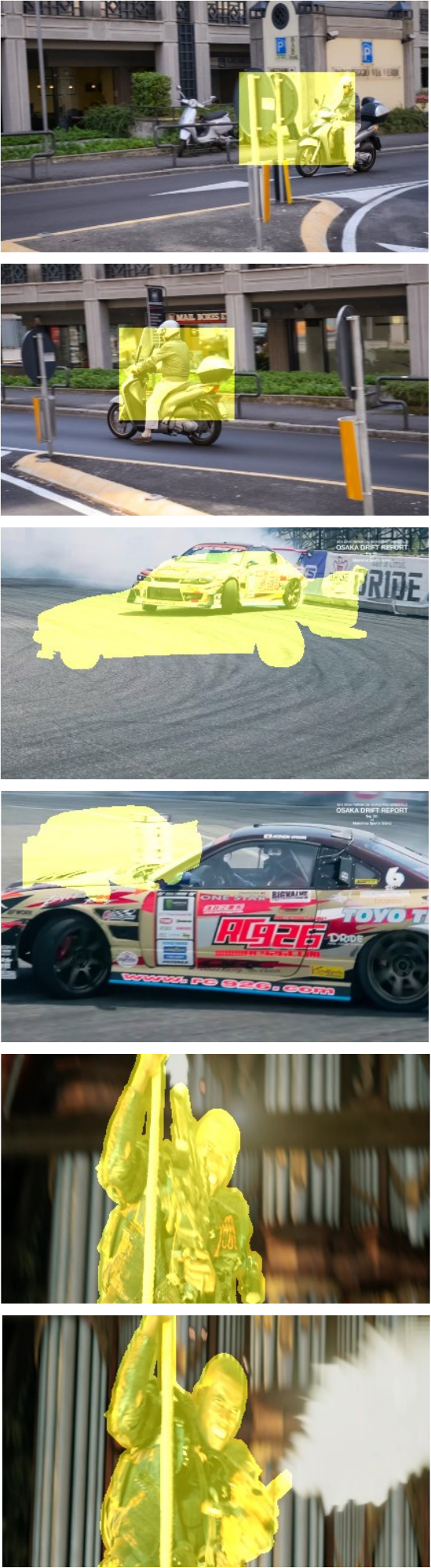}\end{minipage}}
    \subfigure[(b) FFM~\cite{Liu_2021_FuseFormer}]{\begin{minipage}[t]{0.129\linewidth}
        \centering\includegraphics[width=1\linewidth]{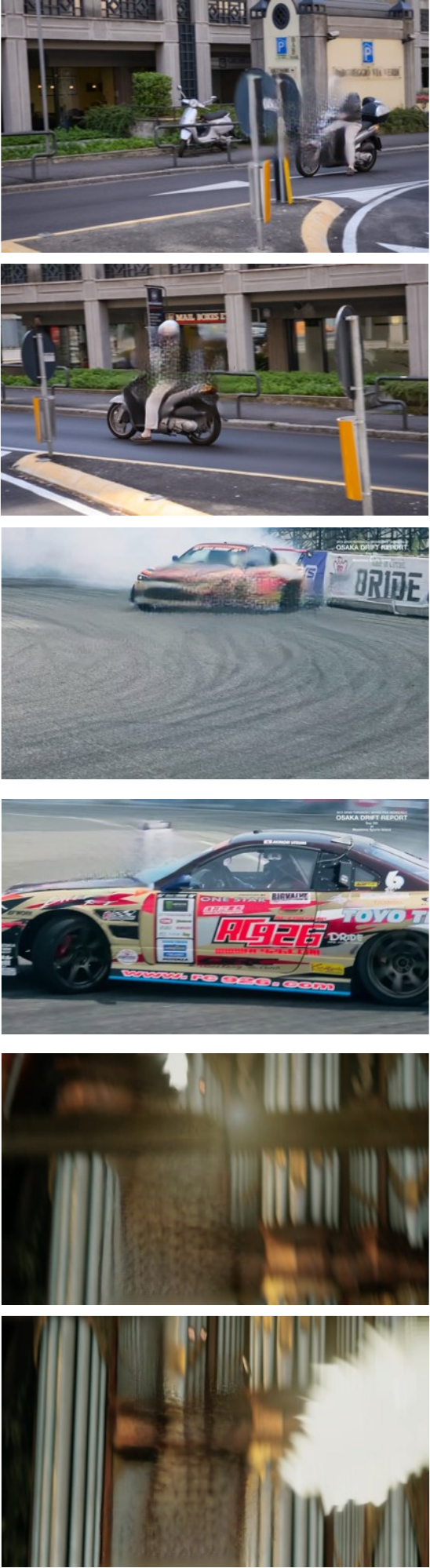}\end{minipage}}
    \subfigure[(c) E2FGVI~\cite{liCvpr22vInpainting}]{\begin{minipage}[t]{0.129\linewidth}
        \centering\includegraphics[width=1\linewidth]{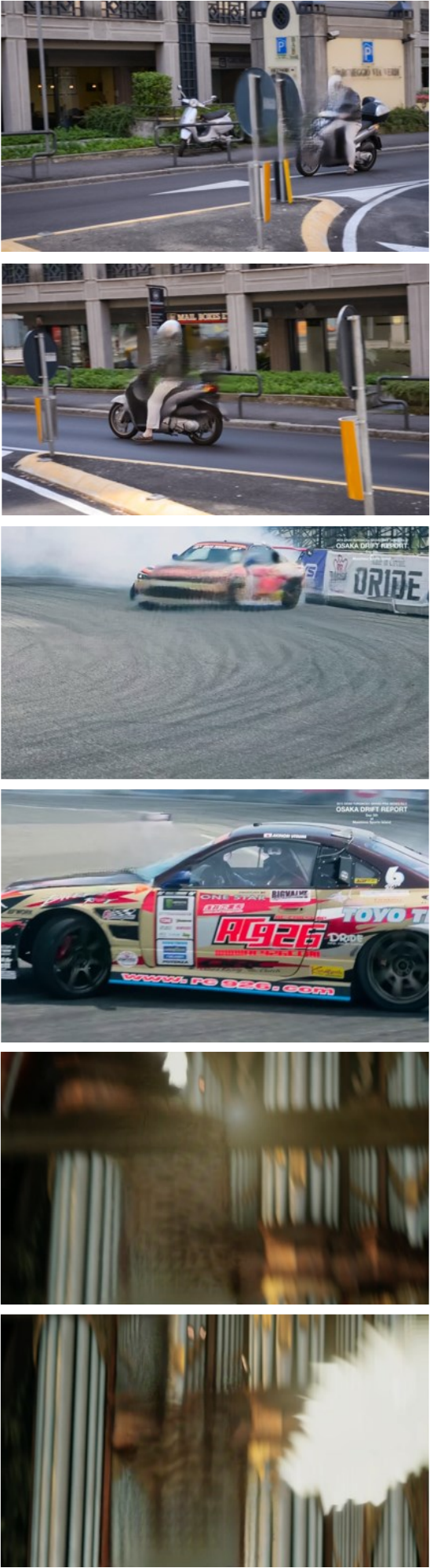}\end{minipage}}
    \subfigure[(d) FGVC~\cite{Gao-ECCV-FGVC}]{\begin{minipage}[t]{0.129\linewidth}
        \centering\includegraphics[width=1\linewidth]{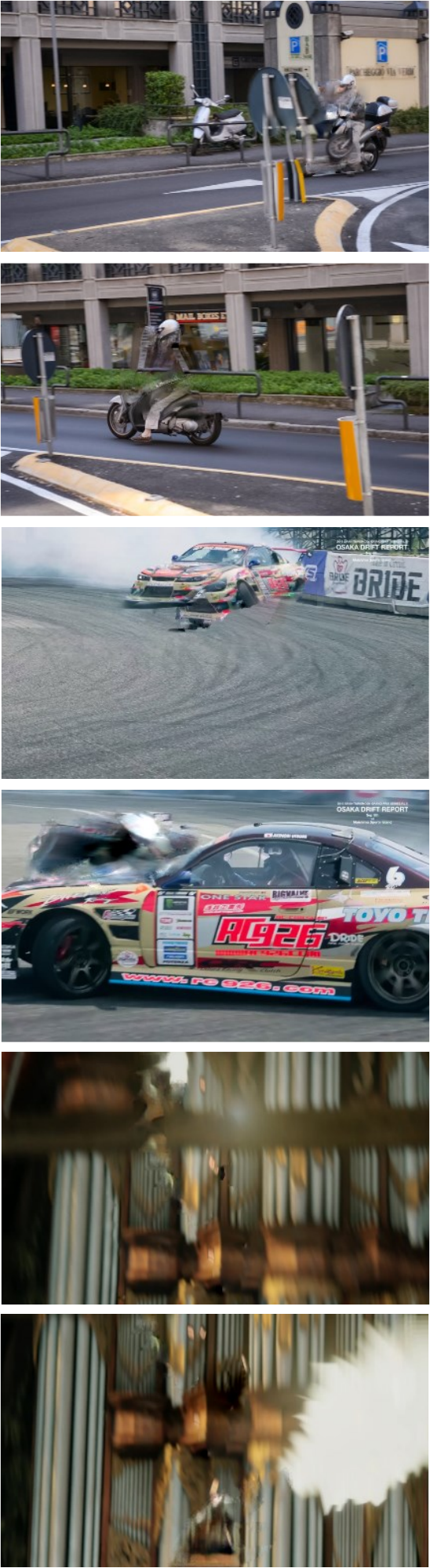}\end{minipage}}
    \subfigure[(e) FGT~\cite{zhang2022flow}]{\begin{minipage}[t]{0.129\linewidth}
        \centering\includegraphics[width=1\linewidth]{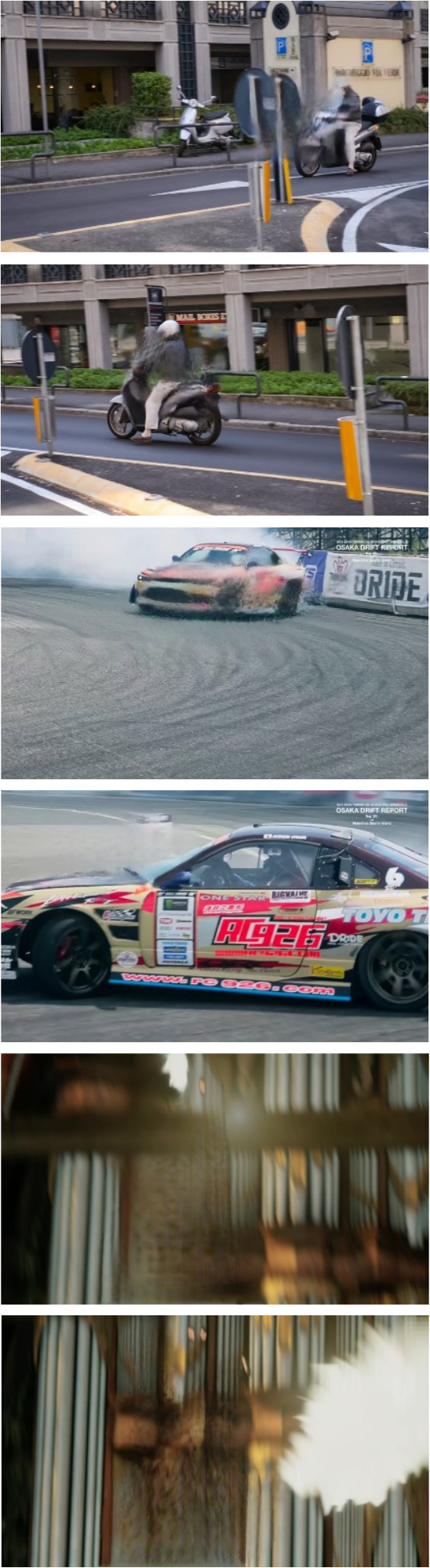}\end{minipage}}
    \subfigure[(f) FGT++]{\begin{minipage}[t]{0.129\linewidth}
        \centering\includegraphics[width=1\linewidth]{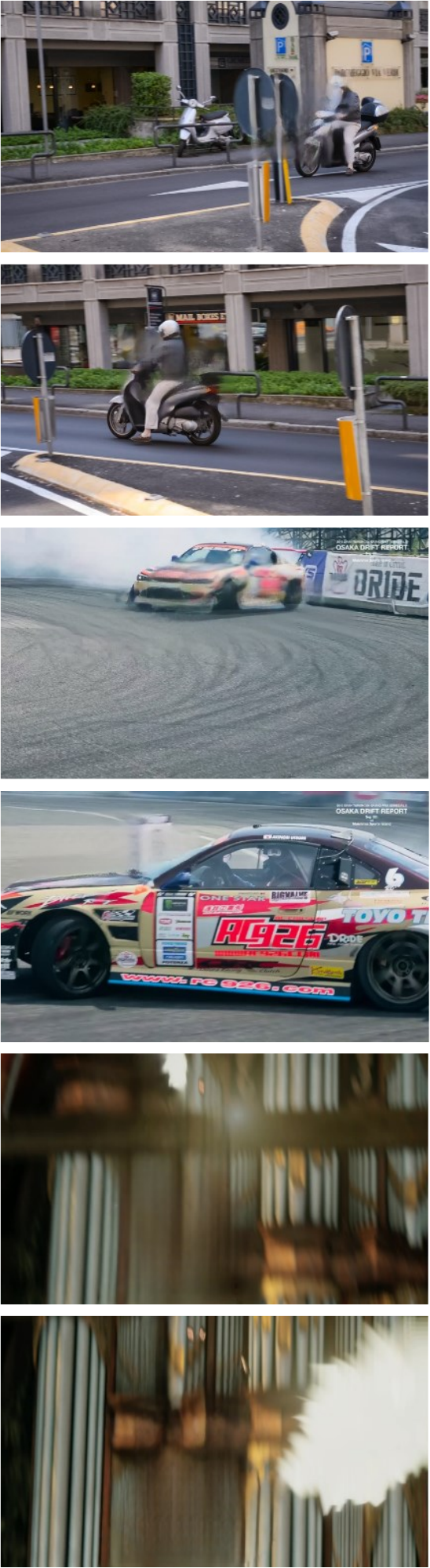}\end{minipage}}
    \subfigure[(g) FGT++*]{\begin{minipage}[t]{0.129\linewidth}
        \centering\includegraphics[width=1\linewidth]{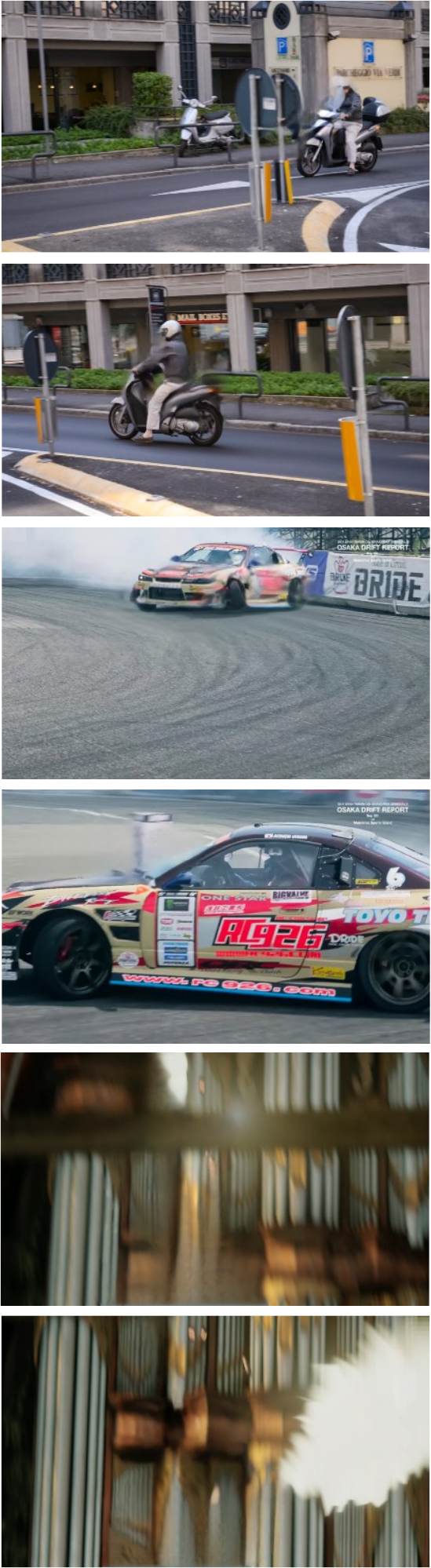}\end{minipage}}
    \captionsetup{font={normalsize}}
    \caption{Qualitative comparison between some recent baselines \cite{Gao-ECCV-FGVC,Liu_2021_FuseFormer,liCvpr22vInpainting,zhang2022flow} and our FGT++. From top to bottom, every two rows display the inpainting results of the square maskset, the object maskset, and the object removal, respectively.}
    \label{quali_rets}
\end{figure*}

\subsection{Qualitative Comparison}
We illustrate the qualitative comparisons between our method and four recent baselines \cite{Gao-ECCV-FGVC,Liu_2021_FuseFormer,liCvpr22vInpainting,zhang2022flow} under the square mask, object mask, and object removal settings in Fig.~\ref{quali_rets}. Compared with these baselines, our method enjoys outstanding visual quality. Thanks to the precise optical flows synthesized by LAFC, FGT++ could restore the corrupted video frames with high fidelity based on the accurate motion trajectory and object clusters formed by the completed optical flows. Moreover, accurate optical flows also play an important role in providing less content propagation error than FGVC \cite{Gao-ECCV-FGVC}, which leads to more precise content propagation results in FGT++*. Therefore, our method can naturally synthesize more visually pleasing video frames.

\begin{figure}[t]
\begin{center}
\includegraphics[width=0.95\linewidth]{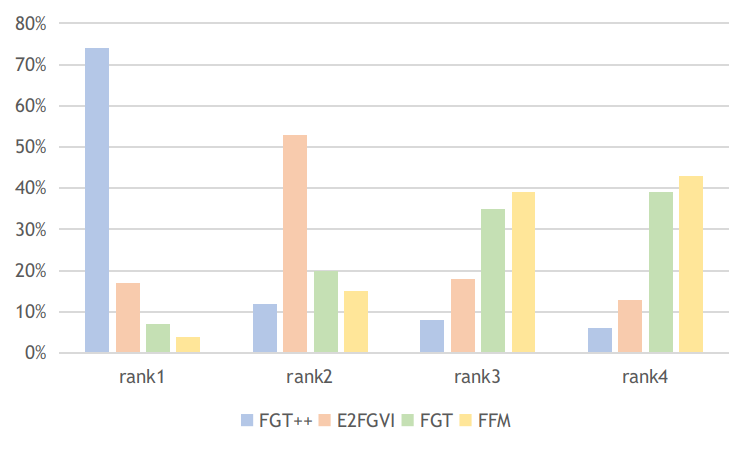}
\end{center}
   \caption{User study results, showing the preference ratios of some highly competitive baselines (FGT~\cite{zhang2022flow}, E2FGVI~\cite{liCvpr22vInpainting}, FFM~\cite{Liu_2021_FuseFormer}) and FGT++. In this study, users are requested to rank the results of the four methods based on perceptual quality.}
\label{user_study}
\end{figure}

We conduct a user study to compare FGT++ with several most competitive baselines, including FGT~\cite{zhang2022flow}, E2FGVI~\cite{liCvpr22vInpainting}, and FFM~\cite{Liu_2021_FuseFormer}. We randomly sample 20 videos from the DAVIS~\cite{caelles20182018} dataset, and recruit 24 volunteers to perform this user study. Specifically, for each video, each volunteer is requested to rank the results of the four methods based on perceptual quality. The results are shown in Fig.~\ref{user_study}. Compared with the baselines, FGT++ is more preferred by the volunteers, which demonstrates the outstanding visual quality of the videos inpainted by FGT++.

\begin{table}[t]
\fontsize{7}{8}\selectfont
\begin{center}
% \scriptsize
\caption{
\textbf{Results of model analysis.}
``FGVC$\rightarrow$FGT++" means that we adopt the completed flows from FGVC to perform the content propagation and to guide the inference of FGT++.
}
\label{tab:ma}
\begin{tabular}{@{}ccccccc@{}}
\toprule
\multirow{2}{*}{Method} & \multicolumn{3}{c}{square} & \multicolumn{3}{c}{object} \\ \cmidrule(r){2-4} \cmidrule(r){5-7}
& PSNR$\uparrow$ & SSIM$\uparrow$ & LPIPS$\downarrow$ & PSNR$\uparrow$ & SSIM$\uparrow$ & LPIPS$\downarrow$ \\ \midrule
FGVC \cite{Gao-ECCV-FGVC} & 32.14 & 0.967 & 0.030 & 33.91 & 0.955 & 0.034 \\
FGVC$\rightarrow$FGT++ & 32.95 & 0.969 & 0.026 & 34.59 & 0.961 & 0.032 \\
FGT++* & \textbf{33.72} & \textbf{0.976} & \textbf{0.022} & \textbf{35.90} & \textbf{0.968} & \textbf{0.027} \\
\bottomrule
\end{tabular}
\end{center}
\end{table}

\begin{table}[t]
\fontsize{8}{9}\selectfont
\begin{center}
% \scriptsize
\caption{
The computational complexity of some baselines and FGT++, assuming the video resolution is 432$\times$256.
}
\label{tab:efficiency}
\begin{tabular}{@{}cccc@{}}
\toprule
Method & Flops & Params & Time \\ \midrule
STTN \cite{yan2020sttn} & 477.91G & 16.56M & 0.22s \\
FFM \cite{Liu_2021_FuseFormer} & 579.82G & 36.59M & 0.30s \\
E2FGVI \cite{liCvpr22vInpainting} & 493.49G & 41.80M & 0.28s \\
FGT \cite{zhang2022flow} & 455.91G & 42.31M & 0.39s \\
FGT++ & 488.59G & 53.30M & 0.53s \\
FGT++* & 488.59G & 53.30M & 2.14s \\
\bottomrule
\end{tabular}
\end{center}
\end{table}

\begin{table}[t]
\fontsize{8}{9}\selectfont
\begin{center}
% \scriptsize
\caption{
Complexity analysis of the newly proposed modules in FGT++.
}
\label{tab:complexity}
\begin{tabular}{@{}ccc@{}}
\toprule
Method & Flops & Params \\ \midrule
TD-MHSA & 1.74G & 1.27M \\
FGFP (encoder) & 21.40G & 4.14M \\
FGFP (in FGF3N) & 1.53G & 0.62M \\
\bottomrule
\end{tabular}
\end{center}
\end{table}

\subsection{Ablation Studies}
\subsubsection{Model analysis} In order to analyze the role that flow completion and transformer-based frame synthesis in video inpainting, we start from FGVC \cite{Gao-ECCV-FGVC} and gradually adapt it with our proposed components. We report the results in Tab.~\ref{tab:ma}. Compared with FGVC, ``FGVC$\rightarrow$FGT++" demonstrates FGT++ is more reasonable than the image inpainting baseline \cite{yu2018generative} to complete the unfilled regions after flow-guided content propagation. Furthermore, the great improvement of ``FGT++*" against the other baselines also demonstrates the importance of the accurate motion trajectory formed by the LAFC completed optical flows in video inpainting. In Tab.~\ref{tab:efficiency}, we compare FGT++ with different transformer-based video inpainting methods. Since Flops in video inpainting are related to the number of frames processed simultaneously, we assume the processed frame number is 20, including 10 local and 10 global frames. This is a common practice in STTN \cite{yan2020sttn} and FFM \cite{Liu_2021_FuseFormer}. Although FGT++ contains more parameters than previous baselines, the computation cost is controllable due to the elaborately designed window partition strategy for temporal and spatial transformer blocks, respectively, which indicates the memory usage of FGT++ is highly competitive. As for the inference speed, if we adopt FGT++ to complete all the missing regions purely, the speed is slower compared with previous transformer baselines but still competitive. If we adopt a flow-guided content propagation procedure (FGT++*), we can obtain much better video inpainting quality, but the speed will degrade to 2.14s/frame because the Poisson blending operation consumes much time.

In Tab.~\ref{tab:complexity}, we provide a detailed complexity analysis of the newly proposed modules in FGT++, including TD-MHSA and FGFP. We measure Flops and parameter size of these modules. Note that the complexity of FGFP in the encoder (before the first transformer block) is different from that in the FGF3N. We distinguish them as ``FGFP (encoder)" and ``FGFP (in FGF3N)" respectively. The analysis shows that the complexity of the newly proposed modules will not bring too much computation burden to FGT++. A theoretical analysis of the computational cost of TD-MHSA against the original large-window MHSA is provided in the supplementary material, which further confirms that TD-MHSA brings only marginal computational cost.

\begin{figure*}[t]
\begin{center}
\includegraphics[width=1\linewidth]{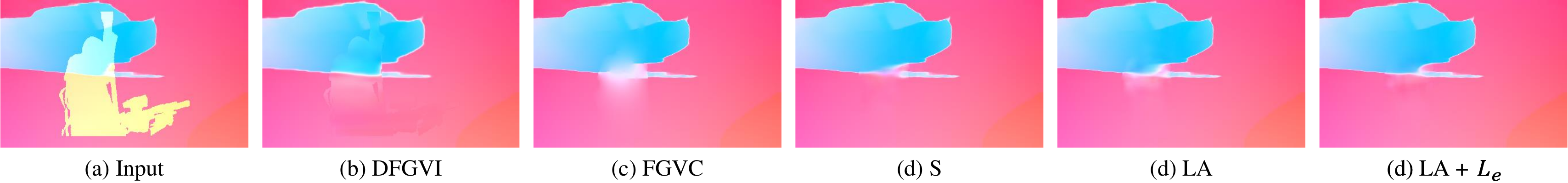}
\end{center}
   \caption{Comparison of the flow completion results of DFGVI \cite{Xu_2019_CVPR}, FGVC \cite{Gao-ECCV-FGVC}, and several variants of our method. S: Single flow completion; LA: Flow completion with local aggregation; $L_e$: Edge loss.}
\label{fig:flowCompare}
\end{figure*}

\begin{table}[t]
\fontsize{8}{9}\selectfont
\begin{center}
\caption{Quantitative results of our proposals for flow completion. S: Single flow completion; LA: Flow completion with local aggregation; $L_e$: Edge loss.}
\label{flow_comp2}
\setlength{\tabcolsep}{0.4mm}{
\begin{tabular}{@{}lccccc@{}}
\toprule
\multirow{2}{*}{Maskset} & \multicolumn{5}{c}{EPE$\downarrow$} \\ \cmidrule(r){2-6}
& DFGVI \cite{Xu_2019_CVPR} & FGVC \cite{Gao-ECCV-FGVC} & S & LA & LA + $L_e$ \\ \midrule
square & 1.161 & 0.633 & 0.546 & 0.524 & \textbf{0.511} \\
object & 1.053 & 0.491 & 0.359 & 0.338 & \textbf{0.328} \\
\bottomrule
\end{tabular}}
\end{center}
\end{table}

\begin{figure}[t]
\begin{center}
\includegraphics[width=0.95\linewidth]{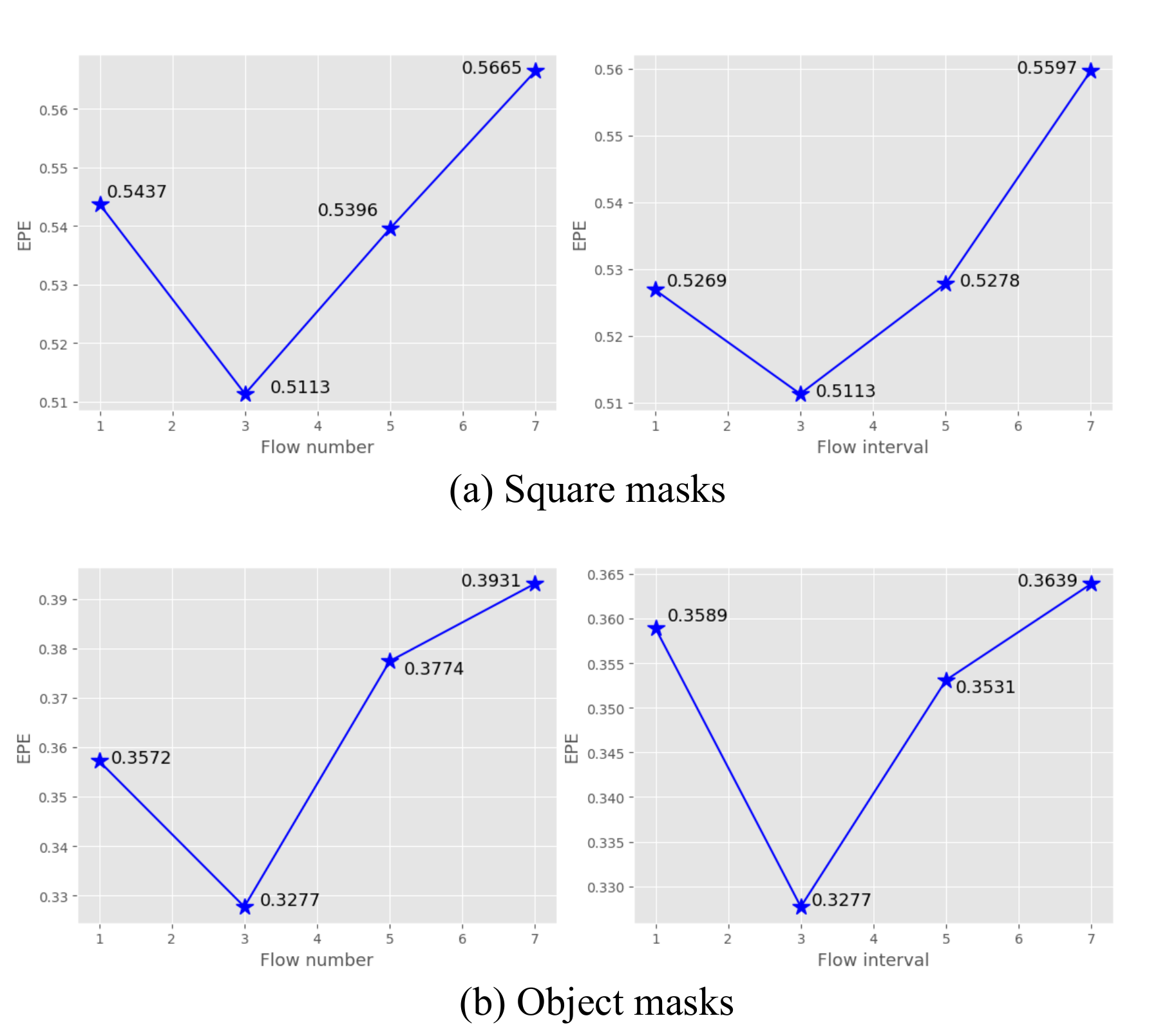}
\end{center}
   \caption{EPE results with varying flow number (when flow interval is 3) or varying flow interval (when flow number is 3) on both square and object masksets.}
\label{LAFC}
\end{figure}

\subsubsection{Flow completion} We report the EPE of LAFC against four baselines, including the previous flow completion methods \cite{Xu_2019_CVPR,Gao-ECCV-FGVC}, LAFC without edge loss supervision and LAFC without local feature aggregation. We show the results in Tab.~\ref{flow_comp2}. With the introduction of local aggregation and edge loss, our method achieves substantial improvement. We illustrate the subjective improvement in Fig.~\ref{fig:flowCompare}. Local aggregation empowers LAFC  to exploit the complementary flow features in a local temporal window, which is beneficial for flow completion under the exposed references. With edge loss, LAFC can synthesize optical flows with clearer motion boundaries. Finally, we report the influence of flow number and flow interval w.r.t. EPE in Fig.~\ref{LAFC}. When the flow number or interval is too small, the target flow cannot utilize abundant references for accurate flow completion, which undermines the performance. However, if the flow number or interval is too large, the flow completion performance will deteriorate gradually because of the relevance degradation of the distant optical flows.

\begin{figure}[t]
\begin{center}
\includegraphics[width=0.95\linewidth]{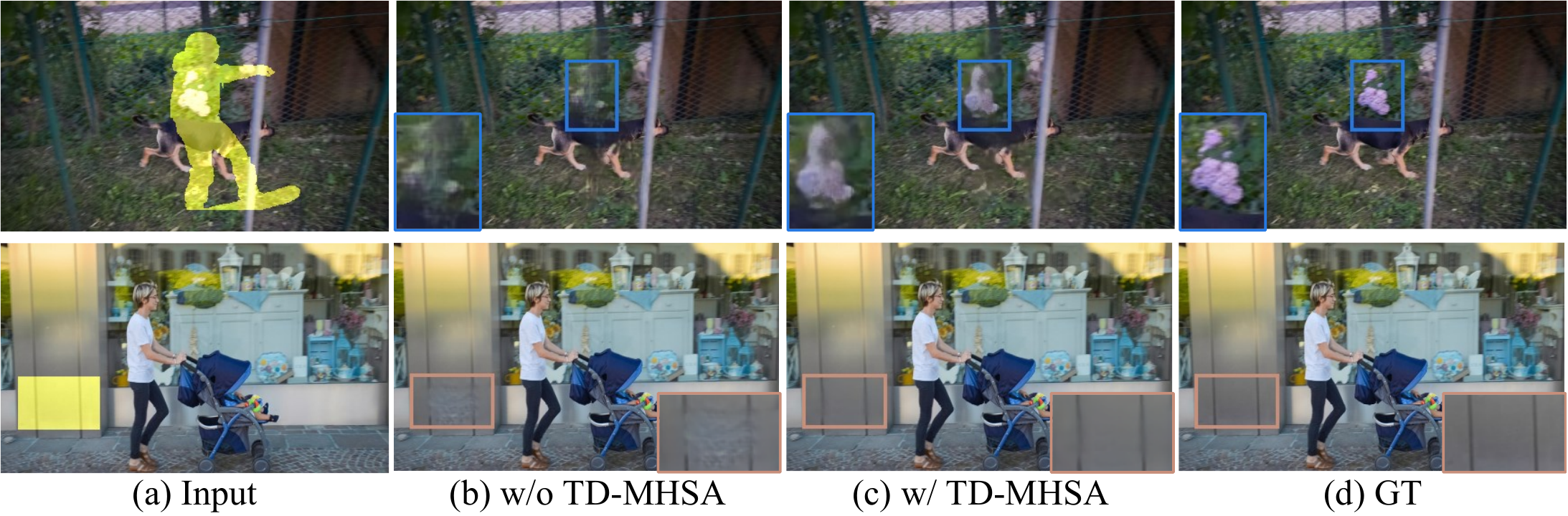}
\end{center}
   \caption{Qualitative comparison for TD-MHSA in the temporal transformer blocks.}
\label{fig:td_mhsa}
\end{figure}

\begin{figure*}[t]
\begin{center}
\includegraphics[width=1\linewidth]{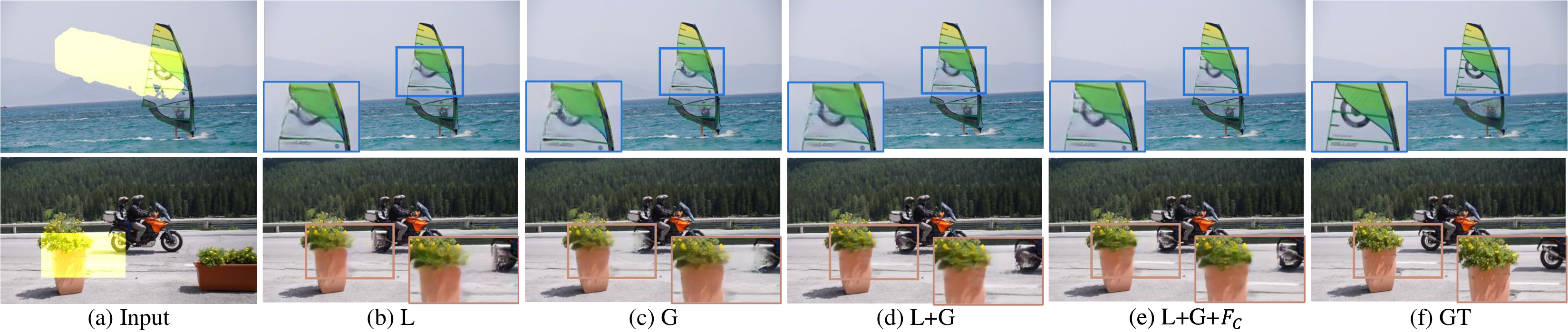}
\end{center}
   \caption{Qualitative comparison for the different components in the spatial transformer blocks. L: Local window attention; G: Global tokens; $F_{c}$: FGFI module.}
\label{fig:transCompare}
\end{figure*}

\begin{table}[t]
\fontsize{8}{9}\selectfont
% \tiny
\begin{center}
\caption{Quantitative results about the different components in FGT++. W: Local window partition; G: Global tokens; F$_{C}$: FGFI module; F$_{P}$: FGFP module; T$_{d}$: TD-MHSA.}
\label{fgdp_trans}
\setlength{\tabcolsep}{1mm}{
\begin{tabular}{@{}lcccccccccc@{}}
\toprule
\multirow{2}{*}{W} & \multirow{2}{*}{G} & \multirow{2}{*}{T$_{d}$} & \multirow{2}{*}{F$_{P}$} & \multirow{2}{*}{F$_{C}$} & \multicolumn{3}{c}{square} & \multicolumn{3}{c}{object} \\ \cmidrule(r){6-8} \cmidrule(r){9-11}
&  &  &  &  &  PSNR$\uparrow$ & SSIM$\uparrow$ & LPIPS$\downarrow$ & PSNR$\uparrow$ & SSIM$\uparrow$ & LPIPS$\downarrow$ \\ \midrule
\checkmark & - & - & - & - & 31.37 & 0.957 & 0.038 & 32.98 & 0.945 & 0.051 \\
- & \checkmark & - & - & - & 31.42 & 0.958 & 0.040 & 33.10 & 0.945 & 0.050 \\
\checkmark & \checkmark & - & - & - & 31.62 & 0.959 & 0.038 & 33.25 & 0.946 & 0.048 \\
\checkmark & \checkmark & \checkmark & - & - & 31.87 & 0.963 & 0.037 & 33.58 & 0.947 & 0.045 \\
\checkmark & \checkmark & - & \checkmark & - & 32.36 & 0.964 & 0.033 & 34.20 & 0.952 & 0.044 \\
\checkmark & \checkmark & - & - & \checkmark & 31.82 & 0.961 & 0.036 & 33.49 & 0.947 & 0.045 \\
\checkmark & \checkmark & \checkmark & \checkmark & - & 32.57 & \textbf{0.965} & \textbf{0.032} & 34.34 & 0.953 & 0.043 \\
\checkmark & \checkmark & \checkmark & \checkmark & \checkmark & \textbf{32.62} & \textbf{0.965} & \textbf{0.032} & \textbf{34.47} & \textbf{0.954} & \textbf{0.042} \\
\bottomrule
\end{tabular}}
\end{center}
\end{table}

\begin{figure}[t]
\begin{center}
\includegraphics[width=1\linewidth]{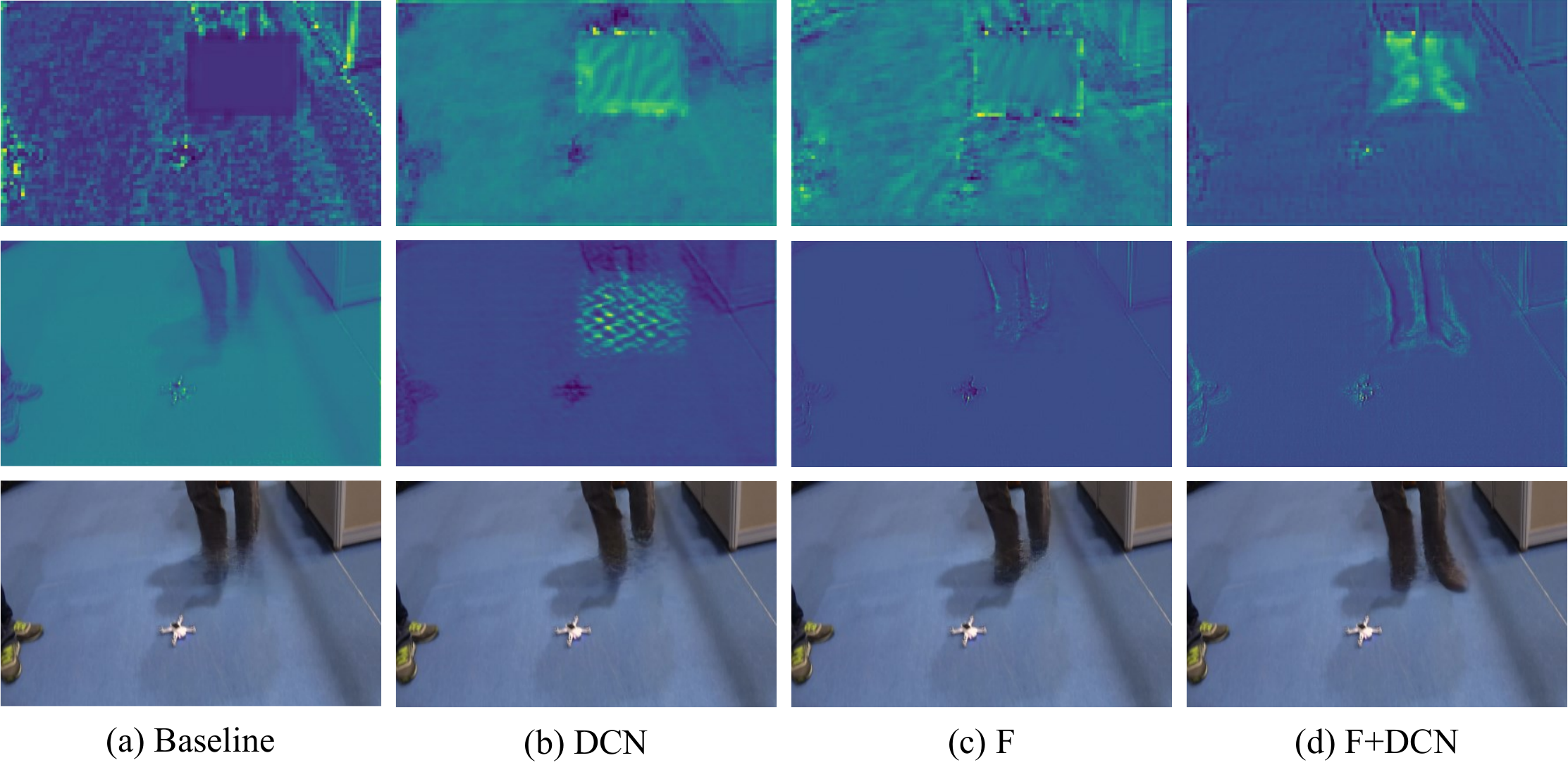}
\end{center}
   \caption{First row: Visualized features processed by FGFP in the encoder. Second row: Visualized features output by the decoder. Third row: Inpainted frames. F: Completed flows; DCN: Deformable convolution.}
\label{fig:feat_vis}
\end{figure}

\begin{figure}[t]
\begin{center}
\includegraphics[width=1\linewidth]{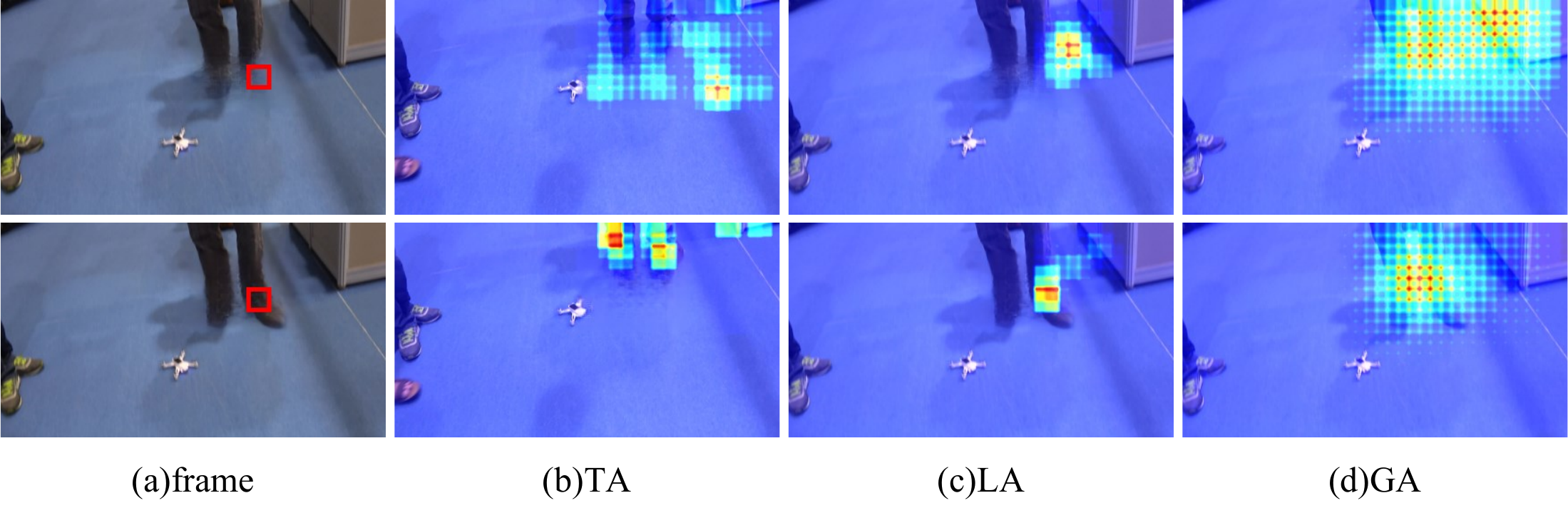}
\end{center}
   \caption{Inpainted frames and visualized attention maps without FGFP (first row) and with FGFP (second row). The red square in (a) indicates the query token for the visualization. TA: Temporal attention; LA: Local attention in the spatial transformer; GA: Global attention in the spatial transformer.}
\label{fig:feat_attn_vis}
\end{figure}

\begin{table}[t]
\fontsize{8}{9}\selectfont
\begin{center}
\caption{Quantitative results about the combination of the flow completion and the deformable convolution in the FGFP module. F: Flow completion; DCN: Deformable convolution.}
\label{flow_prop}
\setlength{\tabcolsep}{0.4mm}{
\begin{tabular}{@{}lcccccc@{}}
\toprule
\multirow{2}{*}{Method} & \multicolumn{3}{c}{square} & \multicolumn{3}{c}{object} \\ \cmidrule(r){2-4} \cmidrule(r){5-7}
& PSNR$\uparrow$ & SSIM$\uparrow$ & LPIPS$\downarrow$ & PSNR$\uparrow$ & SSIM$\uparrow$ & LPIPS$\downarrow$ \\ \midrule
F & 31.76 & 0.960 & 0.036 & 33.55 & 0.946 & 0.047 \\
DCN & 31.55 & 0.959 & 0.037 & 33.41 & 0.945 & 0.048 \\
F+DCN & \textbf{32.36} & \textbf{0.964} & \textbf{0.033} & \textbf{34.20} & \textbf{0.951} & \textbf{0.044} \\
\bottomrule
\end{tabular}}
\end{center}
\end{table}

\begin{figure}[t]
\begin{center}
\includegraphics[width=1\linewidth]{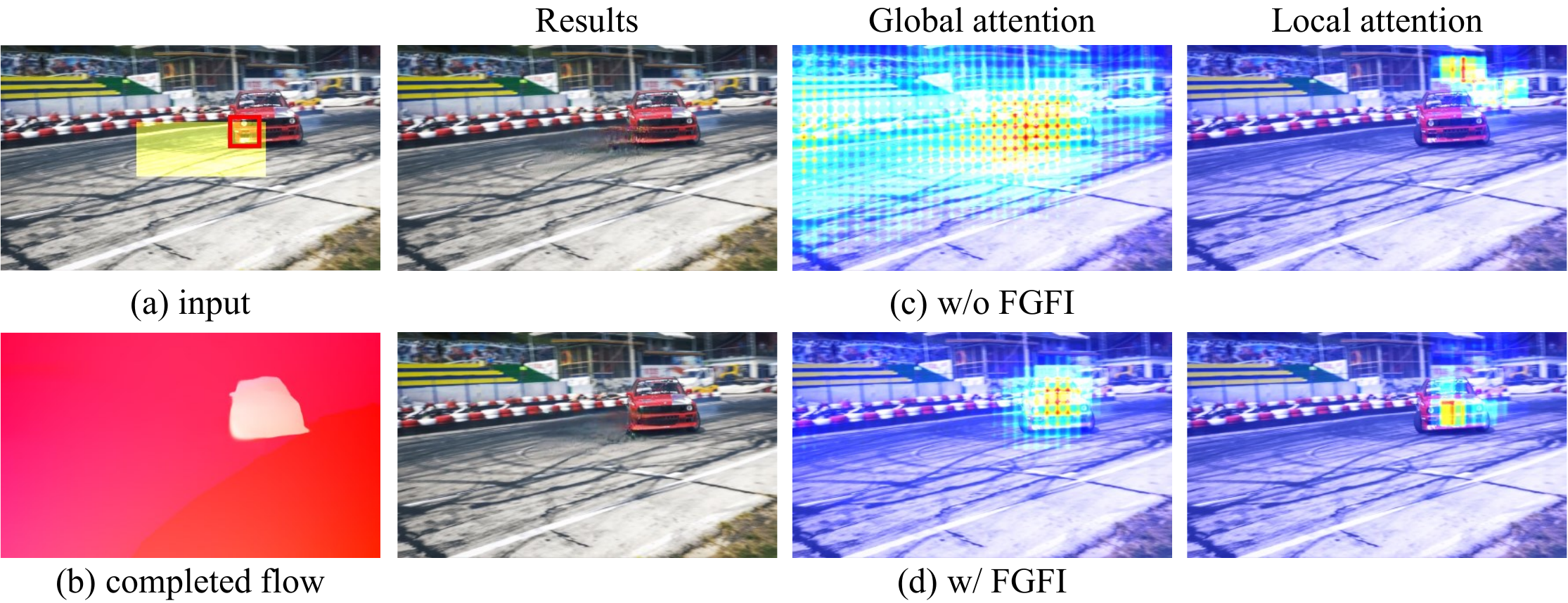}
% [DL] The figure is confusing. I pay some time to understand it. It may be better to revise the figure: put a vertical line between ab and the others; put global attention & local attention left to the final results
\end{center}
   \caption{First column: Input frame and completed flow. Second to fourth columns: Inpainted frames and visualized attention maps without FGFI (first row) and with FGFI (second row). The red square in (a) indicates the query token for the visualization.}
\label{fig:att_vis}
\end{figure}

\subsubsection{Flow-guided transformer} In this part, we adopt FGT++ to synthesize all pixels in the corrupted regions for fair comparisons across different settings. We evaluate the effectiveness of the design of FGT++ from two perspectives. The first is our designed window partition strategy in temporal and spatial transformer blocks, including temporal deformable MHSA (TD-MHSA) and dual perspective spatial MHSA (DP-MHSA). The second is the flow guidance integration module (FGFI) and the flow-guided feature propagation module (FGFP). FGT++ adopts these two components to mitigate the query degradation problem. We report the results in Tab.~\ref{fgdp_trans}. 

Considering the first four rows in Tab.~\ref{fgdp_trans}, we identify the introduction of global tokens in local window-based DP-MHSA leads to a substantial performance boost in FGT++ compared with the existence of only local or global tokens. TD-MHSA also brings significant quantitative improvement to FGT++. We illustrate the qualitative comparisons after introducing specific architecture designs in spatial and temporal transformer blocks in Fig.~\ref{fig:transCompare} and Fig.~\ref{fig:td_mhsa}. We observe the combination of small windows and global tokens leads to the more accurate and smooth structure while maintaining clear object boundaries. In addition, the introduction of TD-MHSA boosts the inpainting quality by integrating motion prior to temporal attention retrieval.

\begin{table}[t]
\fontsize{10}{9}\selectfont
% \small
\begin{center}
\caption{Ablation study for the number of FGFI modules, i.e., the number of spatial transformers from the encoder side integrated with FGFI modules.}
\label{num:fgfi}
\setlength{\tabcolsep}{0.4mm}{
\begin{tabular}{@{}lccccc@{}}
\toprule
\multirow{2}{*}{Maskset} & \multicolumn{5}{c}{PSNR$\uparrow$} \\ \cmidrule(r){2-6}
& 0 & 1 & 2 & 3 & 4 \\ \midrule
square & 31.62 & 31.82 & 31.85 & 31.87 & 31.87 \\
object & 33.25 & 33.49 & 33.51 & 33.52 & 33.52 \\
\bottomrule
\end{tabular}}
\end{center}
\end{table}

\begin{table}[t]
\fontsize{8}{9}\selectfont
\begin{center}
\caption{Quantitative comparison between simple concatenation and FGFI.}
\label{SC}
\setlength{\tabcolsep}{0.4mm}{
\begin{tabular}{@{}lcccccc@{}}
\toprule
\multirow{2}{*}{Method} & \multicolumn{3}{c}{square} & \multicolumn{3}{c}{object} \\ \cmidrule(r){2-4} \cmidrule(r){5-7}
& PSNR$\uparrow$ & SSIM$\uparrow$ & LPIPS$\downarrow$ & PSNR$\uparrow$ & SSIM$\uparrow$ & LPIPS$\downarrow$ \\ \midrule
w/o FGFI & 31.62 & 0.959 & 0.038 & 33.25 & 0.946 & 0.048 \\
Simple concatenation & 31.68 & 0.959 & 0.038 & 33.32 & 0.946 & 0.047 \\
FGFI & \textbf{31.82} & \textbf{0.961} & \textbf{0.036} & \textbf{33.49} & \textbf{0.947} & \textbf{0.045} \\
\bottomrule
\end{tabular}}
\end{center}
\end{table}

The 5th and 6th rows in Tab.~\ref{fgdp_trans} show the effectiveness of our proposed FGFI and FGFP modules. If we combine the proposed modules, the quantitative performance could get boosted further, which demonstrates these modules are complementary to each other. Tab.~\ref{flow_prop} demonstrates the effectiveness of optical flows and learnable deformable offset in feature propagation. Compared with a single component, our method can generate a more accurate motion trajectory, which improves the performance significantly. We visualize the features and the corresponding inpainted frames in Fig.~\ref{fig:feat_vis}. The ``baseline" column represents the model without the FGFP module. With the introduction of flow guidance and learnable deformable offset, we observe the completeness of the structure of features is improved gradually, which is beneficial to the reconstruction of the output features and the inpainted frames (the last two rows). We illustrate the comparison of the attention maps between our method and baseline in Fig.~\ref{fig:feat_attn_vis}. We identify high-quality features that lead to more accurate attention retrieval. Compared with the baseline, our method tends to focus more on the relevant regions, which relieves the query degradation problem.

\begin{table}[t]
\fontsize{8}{9}\selectfont
\begin{center}
\caption{Ablation study for the number of FGFP modules, i.e., the number of transformer groups from the encoder side integrated with FGFP modules (a group is one temporal transformer block and one spatial transformer block). Enc stands for the FGFP module integrated between the encoder and the first transformer block.}
\label{num:fgfp}
\setlength{\tabcolsep}{0.4mm}{
\begin{tabular}{@{}lcccccc@{}}
\toprule
\multirow{2}{*}{Num} & \multicolumn{3}{c}{square} & \multicolumn{3}{c}{object} \\ \cmidrule(r){2-4} \cmidrule(r){5-7}
& PSNR$\uparrow$ & SSIM$\uparrow$ & LPIPS$\downarrow$ & PSNR$\uparrow$ & SSIM$\uparrow$ & LPIPS$\downarrow$ \\ \midrule
Enc & 32.06 & 0.962 & 0.035 & 33.88 & 0.949 & 0.046 \\
Enc+1 & 32.11 & 0.963 & 0.035 & 34.05 & 0.951 & 0.046 \\
Enc+2 & 32.23 & \textbf{0.964} & 0.034 & 34.10 & 0.951 & 0.045 \\
Enc+3 & \textbf{32.36} & \textbf{0.964} & \textbf{0.033} & 34.20 & \textbf{0.952} & \textbf{0.044} \\
Enc+4 & 32.34 & \textbf{0.964} & 0.034 & \textbf{34.21} & 0.951 & \textbf{0.044} \\
\bottomrule
\end{tabular}}
\end{center}
\end{table}

As for the FGFI module, we observe the introduction of the FGFI module in spatial transformer block leads to more accurate object boundaries and a more complete structure, as indicated in Fig.~\ref{fig:transCompare}. We also visualize the optical flows and the corresponding attention map in Fig.~\ref{fig:att_vis}. The red square in Fig.~\ref{fig:att_vis}(a) represents the query token. With flow guidance, FGT++ tends to retrieve the tokens with similar motion patterns (e.g. tokens in car region), which leads to clearer object boundaries for video inpainting with higher fidelity. More visualizations regarding the features, the completed flows, the attention maps, and the video inpainting results are provided in the supplementary material to demonstrate the effectiveness of FGFP and FGFI.

Compared with the simple concatenation, the design of FGFI is more capable of utilizing motion discrepancy for query enhancement, as evidenced in Tab.~\ref{SC}.

\begin{table}[t]
\fontsize{8}{9}\selectfont
\begin{center}
\caption{Ablation study for the number of TD-MHSA, i.e., the number of temporal transformer blocks from the encoder side integrated with TD-MHSA.}
\label{num:td}
\setlength{\tabcolsep}{0.4mm}{
\begin{tabular}{@{}lcccccc@{}}
\toprule
\multirow{2}{*}{Num} & \multicolumn{3}{c}{square} & \multicolumn{3}{c}{object} \\ \cmidrule(r){2-4} \cmidrule(r){5-7}
& PSNR$\uparrow$ & SSIM$\uparrow$ & LPIPS$\downarrow$ & PSNR$\uparrow$ & SSIM$\uparrow$ & LPIPS$\downarrow$ \\ \midrule
1 & 31.66 & 0.960 & 0.038 & 33.34 & 0.946 & 0.048 \\
2 & 31.74 & 0.962 & 0.038 & 33.38 & 0.946 & 0.048 \\
3 & 31.80 & 0.962 & \textbf{0.037} & 33.50 & \textbf{0.947} & 0.046 \\
4 & \textbf{31.87} & \textbf{0.963} & \textbf{0.037} & \textbf{33.58} & \textbf{0.947} & \textbf{0.045} \\
\bottomrule
\end{tabular}}
\end{center}
\end{table}

Besides, we investigate the performance variance w.r.t the number of the FGFI, FGFP and TD-MHSA modules. The baseline is the transformer with large window MHSA in the temporal transformer and DP-MHSA in the spatial transformer, which corresponds to the 3rd row in Tab.~\ref{fgdp_trans}. In Tab.~\ref{num:fgfi}, we observe the FGFI module in the first spatial transformer is crucial, and the others only contribute slight improvement. Such results indicate the motion discrepancy in completed optical flows is helpful for the spatial MHSA, but iterative guidance is not necessary. In contrast, we discover the FGFP module and the TD-MHSA both provide a substantial performance boost to FGT++, as indicated in Tab.~\ref{num:fgfp} and Tab.~\ref{num:td}, respectively. Specifically, the performance of the FGFP module saturates when we impose it on the encoder and the first 6 transformer blocks. The unnecessary FGFP module in the last two transformer blocks indicates that pure feature propagation without further attention retrieval is less helpful in video inpainting.

\begin{figure}[t]
\begin{center}
\includegraphics[width=0.95\linewidth]{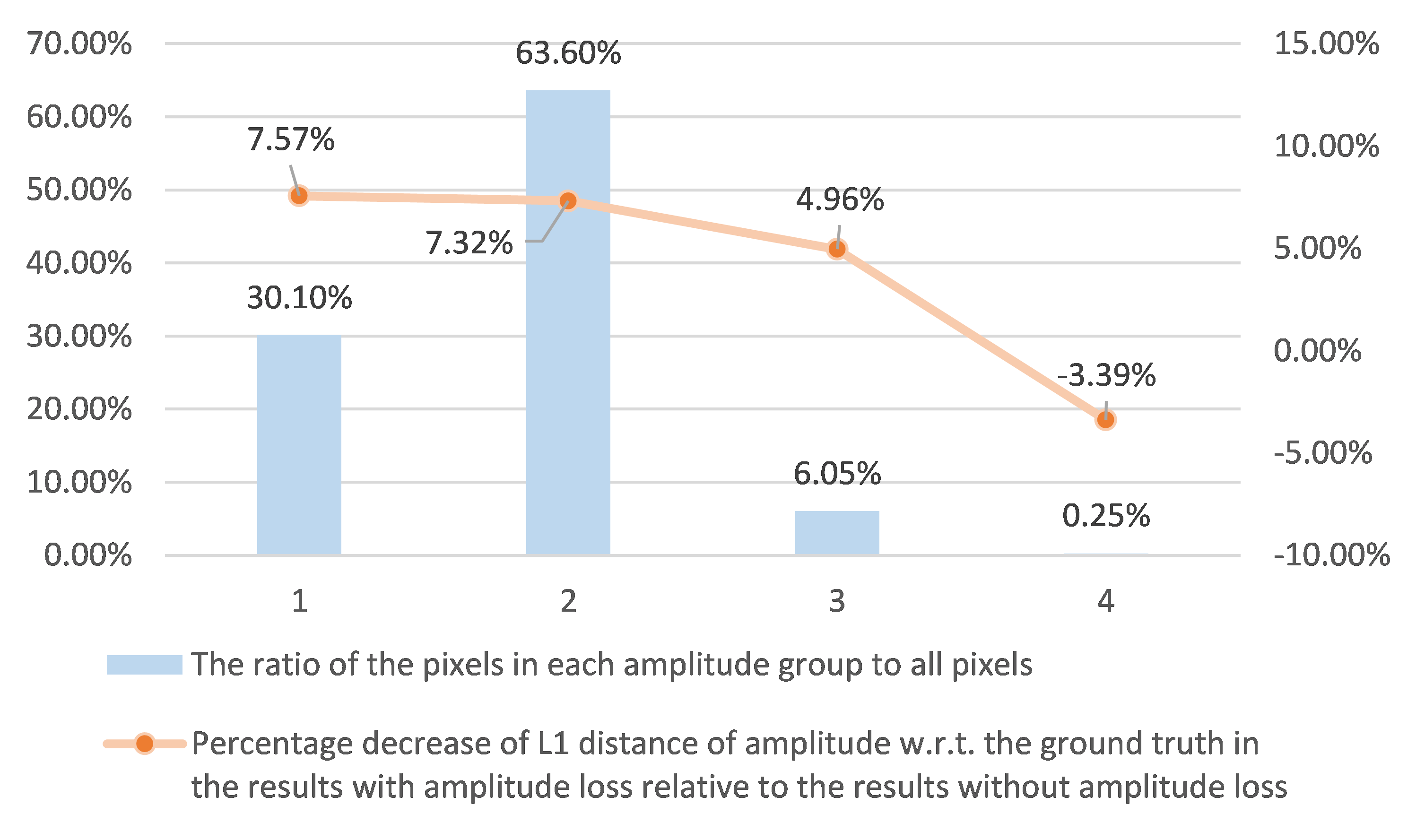}
\end{center}
   \caption{Amplitude-related statistics with respect to each frequency group (the DAVIS dataset with the square maskset setting).}
\label{fig:amp_loss}
\end{figure}

\begin{table}[t]
\fontsize{8}{9}\selectfont
\begin{center}
\caption{Ablation study for the amplitude loss (AMP).}
\label{amp}
\setlength{\tabcolsep}{0.4mm}{
\begin{tabular}{@{}lcccccc@{}}
\toprule
\multirow{2}{*}{Num} & \multicolumn{3}{c}{square} & \multicolumn{3}{c}{object} \\ \cmidrule(r){2-4} \cmidrule(r){5-7}
& PSNR$\uparrow$ & SSIM$\uparrow$ & LPIPS$\downarrow$ & PSNR$\uparrow$ & SSIM$\uparrow$ & LPIPS$\downarrow$ \\ \midrule
w/o AMP & 32.62 & 0.965 & \textbf{0.032} & 34.47 & 0.954 & \textbf{0.042} \\
w. AMP & \textbf{33.01} & \textbf{0.967} & 0.034 & \textbf{34.89} & \textbf{0.956} & 0.045 \\
\bottomrule
\end{tabular}}
\end{center}
\end{table}

\subsubsection{Amplitude loss} We report the results after imposing amplitude loss to FGT++ in Tab.~\ref{amp}. We observe that amplitude supervision could greatly improve the PSNR and SSIM metrics but with a slight sacrifice of LPIPS. We calculate the amplitude map of each frame in the DAVIS dataset, including ground truth and completed frames with or without amplitude loss under the square maskset setting. We divide the amplitude in each frame into four groups. Given a ground truth frame $K$, we transform it to the amplitude domain $\mathcal{A}(K)$ and obtain the maximal amplitude value $\mathcal{A}(K)_{max}$. We define the low-frequency group as group 1 if the amplitude value $\mathcal{A}_i$ satisfies $\lg{\frac{\mathcal{A}_i}{\mathcal{A}(K)_{max}}} \leq -4$; the middle-low frequency group as group 2 if $-4 < \lg{\frac{\mathcal{A}_i}{\mathcal{A}(K)_{max}}} \leq -3$; the middle-frequency group as group 3 if $-3 < \lg{\frac{\mathcal{A}_i}{\mathcal{A}(K)_{max}}} \leq -2$; the high-frequency group as group 4 if $-2 < \lg{\frac{\mathcal{A}_i}{\mathcal{A}(K)_{max}}}$. We illustrate the ratio of the pixels in each amplitude group to all pixels and the percentage decrease of $L_1$ distance of amplitude w.r.t. ground truth in the results with amplitude loss relative to the results without amplitude loss in Fig.~\ref{fig:amp_loss}. We observe that the majority of amplitude maps are low frequency or middle-low frequency, and the $L_1$ distance in these two groups gets the most improvement. Therefore the supervision in the amplitude domain will emphasize the alignment of low-frequency components between the results and the ground truth, which leads to the quantitative performance boost in PSNR and SSIM metrics. However, the high-frequency group occupies only a tiny amount in the amplitude maps, which causes the amplitude loss to ignore the supervision of the high-frequency components in the video inpainting process. Such behavior leads to a slight drop in the LPIPS metric.

\section{Conclusion}
We have proposed FGT++, a transformer-based video inpainting method that exploits the optical flow guidance in several aspects. We have introduced the flow guidance feature integration and flow-guided feature propagation modules to address the query degradation problem. We have designed the temporal deformable MHSA mechanism for the temporal transformer units, and the dual perspective MHSA mechanism for the spatial transformer units. We have also designed a flow completion network to utilize the features of the optical flows in a temporally local window. We have introduced an edge loss for training the flow completion network, and an amplitude loss for training the inpainting network, both of which are shown effective.
Our experimental results have established the effectiveness and efficiency of FGT++. 

The current work has twofold limitations. First, the computational speed of FGT++ is slower than the other transformer-based methods due to the proposed components. Although we have tried to balance performance and speed, the results suggest that we further decrease the computational complexity.
Second, the performance of FGT and FGT++ highly depends on the quality of the completed flows. For a video with large motion, the complete flows may contain severe errors, and FGT and FGT++ may lose effectiveness in exploiting the flow guidance.
We expect that future work may resolve these issues, and may extend the idea of optical flow guidance to other video transformers.

\ifCLASSOPTIONcaptionsoff
  \newpage
\fi



\bibliographystyle{IEEEtran}
\bibliography{ref}

% H.~Kopka and P.~W. Daly, \emph{A Guide to {\LaTeX}}, 3rd~ed.\hskip 1em plus

% \end{thebibliography}

% \end{IEEEbiography}


%\newpage



%\vfill


\end{document}